\documentclass{article}

\usepackage{microtype}
\usepackage{graphicx}
\usepackage{subcaption}
\usepackage{booktabs} 
\usepackage[dvipsnames, table]{xcolor}
\usepackage{soul}
\usepackage{multirow}
\usepackage{hhline}
\usepackage{makecell}
\usepackage{placeins}

\PassOptionsToPackage{hyphens}{url}\usepackage{hyperref}

\usepackage[preprint]{icml2026}

\usepackage{amsmath}
\usepackage{amssymb}
\usepackage{mathtools}
\usepackage{amsthm}
\DeclareMathOperator*{\argmin}{argmin}
\newcommand{\norm}[1]{\left\lVert#1\right\rVert}
\DeclarePairedDelimiter{\ceil}{\lceil}{\rceil}

\usepackage[capitalize,noabbrev]{cleveref}

\theoremstyle{plain}

\theoremstyle{definition}

\theoremstyle{remark}

\usepackage[textsize=tiny]{todonotes}

\icmltitlerunning{RaZeR: Pushing the Limits of NVFP4 Quantization with Redundant Zero Remapping}

\usepackage{multirow, tabularx, threeparttable} 
\usepackage{arydshln}
\usepackage{makecell}

\newcommand{\workname}{\text{RaZeR}}

\usepackage{tikz}
\newcommand*\circled[1]{\tikz[baseline=(char.base)]{\node[shape=circle,fill,inner sep=0.25pt] (char) {\textcolor{white}{#1}};}}

\begin{document}

\twocolumn[
  \icmltitle{RaZeR: Pushing the Limits of NVFP4 Quantization with \\Redundant Zero Remapping
}

  \icmlsetsymbol{equal}{*}

  \vspace{-5pt}
  \begin{icmlauthorlist}
    \icmlauthor{Yuzong Chen}{equal,Cornell}
    \icmlauthor{Xilai Dai}{equal,Cornell}
    \icmlauthor{Jake Hyun}{equal,CornellCS}
    \icmlauthor{Chi-Chih Chang}{Cornell}
    \icmlauthor{Wonsuk Jang}{Stanford}
    \icmlauthor{Yuheng Wu}{Stanford} \\
    \icmlauthor{Thierry Tambe}{Stanford}
    \icmlauthor{Jae-sun Seo}{Cornell}
    \icmlauthor{Mohamed S. Abdelfattah}{Cornell}
  \end{icmlauthorlist}

  \icmlaffiliation{Cornell}{Department of Electrical and Computer Engineering, Cornell Tech, NY, USA}
  \icmlaffiliation{CornellCS}{Department of Computer Science, Cornell Tech, NY, USA}
  \icmlaffiliation{Stanford}{Department of Electrical Engineering, Stanford University, CA, USA}

  \icmlcorrespondingauthor{Mohamed S. Abdelfattah}{mohamed@cornell.edu}

  \icmlkeywords{Machine Learning, ICML}

  \vskip 0.25in
]



\printAffiliationsAndNotice{$^*$Equal contribution. The order of three authors is alphabetical by last name.}  

\begin{abstract}
  The recently introduced NVFP4 format demonstrates remarkable performance and memory benefits for quantized large language model (LLM) inference. However, we observe two types of redundancy in NVFP4 encoding: 
  (1) The FP4 element format naturally exposes an unused quantization value due to its sign-magnitude representation that contains both positive and negative zeros. 
  (2) The FP8 block scaling factor has an unused sign bit because it is always positive. 
  Additionally, we find that LLM weights are more tolerant to a lower-precision block scaling factor. 
  Based on these observations, we propose \underline{R}edund\underline{a}nt \underline{Ze}ro \underline{R}emapping (\workname{}), an enhanced numerical format that pushes the limits of NVFP4 for more accurate LLM quantization under the same memory footprint. \workname{} leverages the redundant bits of the block scaling factor to adaptively remap the redundant FP4 zero to additional quantization values with improved accuracy. To demonstrate the practicality of~\workname{}, we design efficient GPU kernels for \workname{}-quantized LLM inference and propose novel hardware to natively support this. Extensive experiments validate \workname{}'s superior performance for 4-bit LLM quantization. For example, relative to native NVFP4, \workname{} reduces the average perplexity loss by $34.6\%$ and $31.2\%$ under weight-only and weight-activation quantization, respectively. Code is available at: \url{https://github.com/yc2367/NVFP4-RaZeR}.
\end{abstract}

\section{Introduction} \label{sec:intro}

\begin{figure}[t]
  \vskip 0.1in
  \begin{center}
    \centerline{\includegraphics[width=\columnwidth]{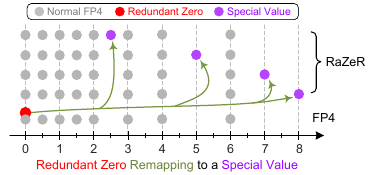}}
    \caption{
      \textbf{\workname{} overview}. The negative set of normal FP4 values are omitted for simplicity. 
    }
    \label{fig:razer_overview}
  \end{center}
  \vspace{-20pt}
\end{figure}

The substantial memory bandwidth and computational demands of large language models (LLMs) present critical challenges for their deployment. To tackle this, quantization has emerged as a promising solution for efficient LLM inference. By reducing the operand bit-width, quantization not only decreases the memory footprint, but also improves the computational efficiency on low-precision hardware~\cite{quant_survey, int_vs_fp}. 
While many prior approaches use the standard integer format to represent quantized values~\cite{atom, awq, gptq, omniquant, duquant, smoothquant}, recent literature has started to focus on custom low-precision formats that better adapt to LLM tensor distributions, thus reducing the quantization error~\cite{microscaling, nvfp4, nf4, students, blockdialect}. 
Among these, the custom 4-bit NVFP4 format~\cite{nvfp4}, supported by NVIDIA Blackwell GPUs, delivers superior performance for LLM quantization, outperforming earlier 4-bit formats such as INT4 and MXFP4~\cite{fp_quant, int_vs_fp}.
On top of FP4, NVFP4 employs block-wise quantization, in which each group of 16 values shares an FP8 scaling factor, with an additional FP32 scaling factor applied per tensor. This fine-grained scaling enables a larger range of tensor values to be represented in FP4 precision while lowering error over prior data formats.

Despite the performance benefits of NVFP4, adopting it for accurate 4-bit weight-only and weight-activation quantization remains challenging due to the limited set of quantization values offered by FP4.
In this paper, we propose \underline{R}edund\underline{a}nt \underline{Ze}ro \underline{R}emapping (\workname{}), an enhanced NVFP4 format, illustrated in Fig.~\ref{fig:razer_overview}. 
FP4 can represent both positive and negative zeros, of which one is redundant. \workname{} adaptively remaps the redundant zero to an additional quantization level, referred to as the \textit{special value}. 
By carefully selecting this special value, each NVFP4 block can be quantized with the basic FP4 values and a useful special value, thereby reducing per-block quantization error. 
Additionally, \workname{} exploits redundancy in the block scaling factor to encode special value metadata, thus maintaining the same memory footprint as the baseline NVFP4 format while achieving much higher accuracy. 

Beyond algorithmic innovation, we thoroughly study the hardware implications of \workname{}. First, we demonstrate the practicality of deploying \workname{} on current GPU platforms by implementing high-performance kernels to support weight-only quantized LLMs using \workname{} on Nvidia Blackwell GPUs.  Second, we design a modified NVFP4 tensor core that efficiently performs multiply-accumulate (MAC) operations when weights and activations are quantized in \workname{}. 
Our main contributions are summarized as follows:
\begin{itemize}
    \item We propose \workname{}, an enhanced NVFP4 format that exploits redundancy in NVFP4 block scaling to repurpose the redundant FP4 zero as additional quantization values, enabling more accurate LLM quantization at the same memory footprint. 
    \item We implement efficient GPU kernels that enable \workname{} to be used for 4-bit weight-only quantized LLM inference on Blackwell GPUs, achieving performance comparable to or better than prior kernels. 
    \item We propose a modified NVFP4 tensor core that adds native support for \workname{} with tiny silicon overhead.
    \item Through comprehensive evaluation, we show that \workname{} outperforms existing 4-bit methods across both weight-only and weight-activation quantization.
\end{itemize}
%

\section{Related Work} \label{sec:related_work}

\subsection{Quantization Formats}
Traditionally, a majority of LLM quantization algorithms rely on the standard integer format~\cite{awq, gptq, smoothquant, atom, omniquant, duquant}. However, LLMs often exhibit outliers that substantially expand the dynamic range, leading to large error under integer quantization. To address this issue, recent work has explored custom quantization formats that better match the distribution of LLM tensors. At 8 bits, ZeroQuant-FP~\cite{zeroq-fp} demonstrates that FP8~\cite{fp8} significantly outperforms INT8 for activation quantization, achieving near-lossless accuracy when combined with INT8 weights. At 4 bits, lookup-based formats such as Normal-Float~\cite{nf4} and Student-Float~\cite{students} leverage statistical quantile functions to define the 16 representable values. However, these methods typically store the lookup table in FP16, necessitating high-precision MAC operations with significant energy overhead. In contrast, the FP4-E2M1 format offers a favorable trade-off between model accuracy and hardware efficiency~\cite{llm-fp4, int_vs_fp}, resulting in growing industry adoption and practical impact~\cite{mx-format, nvfp4}. Our work further improves FP4 quantization by remapping its redundant zero to a carefully chosen special values, achieving better model accuracy while being highly compatible with FP4 low-precision hardware.

\subsection{Block-wise Quantization}
Block-wise quantization is a widely adopted technique to reduce the accuracy loss of quantized models~\cite{block-quant, vs-quant}.
Prior software-based  solutions~\cite{atom, awq, qserve} typically rely on high-precision block scaling factors and slow CUDA cores to perform runtime dequantization.
On the other hand, the recent Blackwell GPU natively supports various block-wise quantization formats such as NVFP4~\cite{nvfp4}, enabling hardware-accelerated scaling under small block sizes (e.g., 16) with improved accuracy. 

Motivated by this industry trend, several works have introduced modifications to the NVFP4 quantization algorithm. MR-GPTQ \cite{fp_quant} applies GPTQ~\cite{gptq} and Hadamard rotation on top of NVFP4, but finds that adding Hadamard rotation generally hurts model accuracy. FourOverSix~\cite{4over6} improves NVFP4 with adaptive blocking scaling, where instead of scaling all blocks to the full range of FP4, some blocks can be scaled to a narrower range with less error.  
Our work follows this line of research and investigates strategies to push the limits of NVFP4 to a new state-of-the-art.
Unlike prior works that focus on refining the NVFP4 \emph{quantization algorithm}, \workname{} aims to enhance the NVFP4 \emph{format} by exploiting redundancy in its block scaling factor to encode additional special quantization values. 
Furthermore, we design high-performance GPU kernels and efficient tensor core hardware to demonstrate the practicality of adopting \workname{} on both current and future GPU platforms.

\section{Preliminaries on NVFP4} \label{sec:nvfp4}

\begin{figure}[t]
  \begin{center}
    \centerline{\includegraphics[width=\columnwidth]{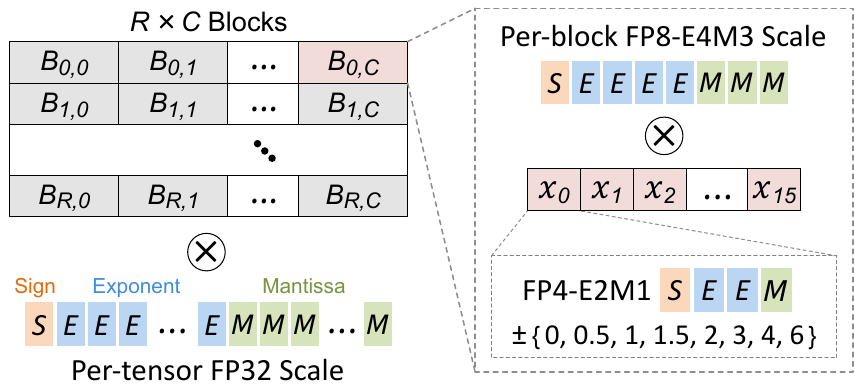}}
    \caption{
      \textbf{NVFP4 overview}. A 16-value block of FP4 elements share an FP8-E4M3 scale, followed by a tensor-wise FP32 scale. 
    }
    \label{fig:nvfp4_overview}
  \end{center}
  \vspace{-15pt}
\end{figure}

Fig.~\ref{fig:nvfp4_overview} shows the high-level overview of the NVFP4 format. A tensor is divided into $R \times C$ blocks that share a tensor-wise FP32 scale, and each block contains 16 quantized FP4 values $\left[x_0\,, x_1\,, ...\,, x_{15}\right]$ that share a FP8-E4M3\footnote{In this paper, we use E$x$M$y$ to represent a format with $x$-bit exponent and $y$-bit mantissa.} scale. Mathematically, the NVFP4 quantization process can be expressed as follows:
\begin{equation}
    \Delta^{\text{FP32}} = \frac{ |X^{\text{FP32}}|_{\text{max}} }{ Q^{\text{FP8}}_{\text{max}} \cdot Q^{\text{FP4}}_{\text{max}}} 
    \label{eq:fp32_scale}
\end{equation} 
\begin{equation}
    \Delta^{\text{FP8}}_{i} = \left\lfloor 
        \frac{ |X^{\text{FP32}}_{i}|_{\text{max}} }{ \Delta^{\text{FP32}} \cdot Q^{\text{FP4}}_{\text{max}}} 
        \,,\, Q^{\text{FP8}}
    \right\rceil
    \label{eq:fp8_scale}
\end{equation} 
\begin{equation} 
    \bar{X}^{\text{FP4}}_{i} = \left\lfloor
        \frac{ X^{\text{FP32}}_{i} }{ \Delta^{\text{FP32}} \cdot \Delta^{\text{FP8}}_{i}} 
        \,,\, Q^{\text{FP4}}
    \right\rceil
    \label{eq:fp4_value_quant}
\end{equation} 
where $X^{\text{FP32}}$ is the high-precision tensor, $\bar{X}^{\text{FP4}}$ is the quantized FP4 tensor, $Q^{\text{FP8}}$ and $Q^{\text{FP4}}$ are representable quantization values of FP8 and FP4, $\Delta^{\text{FP32}}$ is the tensor-wise FP32 scaling factor, $\Delta^{\text{FP8}}_{i}$ is the FP8 scaling factor of block $i$, and $\left\lfloor \,\cdot\, \right\rceil$ is the rounding function that casts a high-precision FP32 value to a low-precision FP8 or FP4 value.

\comment{
\begin{equation}
\hat{X}_i
=
\Delta^{\mathrm{FP32}}
\cdot
\Delta_i^{\mathrm{FP8}}
\cdot
\bar{X}_i^{\mathrm{FP4}}
\end{equation}
}

The rounding function inevitably introduces discrepancy between the original and quantized operands. Consequently, there are two types of quantization error in NVFP4: (1) Rounding of block scaling factors to FP8; (2) Rounding of individual element values to FP4. Because FP8 offers much higher precision than FP4, reducing the error caused by FP4 rounding is more effective for improving the overall accuracy of NVFP4, which is the focus of \workname{}.

\section{Methodology} \label{sec:methodology}

In this section, we start by describing two sources of redundancy in the current NVFP4 encoding. We then leverage these redundancies to derive \workname{}, which increases NVFP4 accuracy with no memory overhead. 

\subsection{Redundancy in NVFP4 Block Scale Encoding} 
\label{sec:nvfp4_scale_redun}
As described in Section~\ref{sec:nvfp4}, NVFP4 quantizes the block scale and the block element to FP8 and FP4, respectively. We observe that both FP8 and FP4 formats have redundancy under the context of NVFP4. According to the standard of \citet{mx-format}, the FP8-E4M3 and FP4-E2M1 formats can be expressed as follows:
\begin{equation}
    Q^{\text{FP8}} =
    \begin{cases}
        (-1)^{S} \cdot 2^{E-7} \cdot \left(1+\frac{M}{8}\right) & \text{if $E \neq 0$} \\[0.5ex]
        (-1)^{S} \cdot 2^{-6} \cdot \frac{M}{8} & \text{if $E = 0$}
    \end{cases}
    \label{eq:fp8_value}
\end{equation} 
\begin{equation}
    Q^{\text{FP4}} =
    \begin{cases}
        (-1)^{S} \cdot 2^{E-1} \cdot \left(1+\frac{M}{2}\right) & \text{if $E \neq 0$} \\[0.5ex]
        (-1)^{S} \cdot \frac{M}{2} & \text{if $E = 0$}
    \end{cases}
    \label{eq:fp4_value}
\end{equation} 
where, $S$, $E$, $M$ are sign, exponent, and mantissa of the quantized value, respectively. Based on Eq.~\ref{eq:fp8_value}, the FP8 format contains a sign bit to encode both positive and negative values. However, in Eq.~\ref{eq:fp32_scale} and Eq.~\ref{eq:fp8_scale}, both numerator and denominator values are positive, resulting in  $\Delta^{\text{FP32}}$ and $\Delta^{\text{FP8}}_{i}$ values that are always positive. Thus, the FP8 block scaling factor used in NVFP4 natively contains a redundant sign bit, resulting in an effective width of 7 bits. 

\begin{table} [t]
  \centering
  \setlength{\tabcolsep}{3pt}
  \renewcommand{\arraystretch}{1.15}
  \footnotesize
  \caption{Wikitext-2 and C4 perplexity ($\downarrow$) using different block scale formats for weight-only NVFP4 quantization. The scale format E3M3 has no accuracy loss compared to the baseline E4M3 format. Results for other models are available in Appendix~\ref{apdx:A}.}
  \vspace{-5pt}
    \begin{tabular}{ c | c  c | c  c | c  c | c  c }
      \toprule
        & \multicolumn{4}{c|}{\textbf{Llama}} & \multicolumn{4}{c}{\textbf{Qwen}} \\
      \midrule
        & \multicolumn{2}{c|}{3.1-8B} & \multicolumn{2}{c|}{3.2-3B} & \multicolumn{2}{c|}{3-4B} & \multicolumn{2}{c}{3-8B} \\
        & Wiki & C4 & Wiki & C4 & Wiki & C4 & Wiki & C4 \\
      \midrule
        E4M3 & 6.63 & 9.48 & 8.24 & 11.10 & 13.63 & 16.85 & 9.92 & 13.55 \\
      \midrule
        E4M2 & 6.67 & 9.53 & 8.29 & 11.17 & 13.86 & 16.89 & 9.99 & 13.59 \\
        \textbf{E3M3} & \textbf{6.63} & \textbf{9.48} & \textbf{8.24} & \textbf{11.10} & \textbf{13.63} & \textbf{16.85} & \textbf{9.92} & \textbf{13.55} \\
        E2M4 & 6.64 & 9.50 & 8.25 & 11.11 & 14.13 & 16.97 & 10.11 & 13.65 \\
        E3M2 & 6.67 & 9.53 & 8.29 & 11.17 & 13.96 & 16.90 & 9.99 & 13.59 \\
        E2M3 & 6.72 & 9.63 & 8.32 & 11.21 & 14.22 & 17.22 & 10.22 & 13.79 
      \\
      \bottomrule
    \end{tabular}
  \label{tab:w_scale_format}
\end{table}

Motivated by this observation, we explore the possibility of further reducing the bit width of block scaling factors. Table~\ref{tab:w_scale_format} shows the perplexity of Wikitext-2~\cite{wikitext} and C4~\cite{c4} datasets across four models from Llama-3~\cite{llama-3} and Qwen-3~\cite{qwen3} families, when applying different block scale formats to NVFP4 weight-only quantization. We find that the E3M3 format has no accuracy drop compared to the baseline E4M3 block scale, while reducing the encoding cost to 6 bits. This can be attributed to the small dynamic range of LLM weights, as observed in prior study~\cite{smoothquant}. Since the number of exponent bits controls the numerical range, weights can tolerate a scale format with fewer exponent bits. Accordingly, we adopt the E3M3 format for weight block scaling.

We also quantify the impact of different block scale formats on NVFP4 activation quantization, as shown in Table~\ref{tab:a_scale_format}.
We observe that activations are substantially less tolerant to reduced-precision block scales, which can be attributed to their large dynamic range with extreme outliers~\cite{llm-int8}. This explains why the E4M2 format performs closest to the baseline E4M3 scale. Nevertheless, E4M2 still exhibits noticeable accuracy degradation. Therefore, we choose to retain the 7-bit E4M3 scale format for NVFP4 activation quantization, which offers 1 free bit per block.

\begin{table} [t]
  \centering
  \setlength{\tabcolsep}{2.5pt}
  \renewcommand{\arraystretch}{1.15}
  \footnotesize
  \caption{Wikitext-2 and C4 perplexity ($\downarrow$) using different block scale formats for activation-only NVFP4 quantization. Results for other models are available in Appendix~\ref{apdx:A}.}
  \vspace{-5pt}
    \begin{tabular}{ c | c  c | c  c | c  c | c  c }
      \toprule
        & \multicolumn{4}{c|}{\textbf{Llama}} & \multicolumn{4}{c}{\textbf{Qwen}} \\
      \midrule
        & \multicolumn{2}{c|}{3.1-8B} & \multicolumn{2}{c|}{3.2-3B} & \multicolumn{2}{c|}{3-4B} & \multicolumn{2}{c}{3-8B} \\
        & Wiki & C4 & Wiki & C4 & Wiki & C4 & Wiki & C4 \\
      \midrule
        E4M3 & 6.53 & 9.34 & 8.12 & 10.89 & 13.81 & 16.93 & 9.84 & 13.49 \\
      \midrule
        \textbf{E4M2} & \textbf{6.59} & \textbf{9.43} & \textbf{8.18} & \textbf{10.98} & \textbf{13.94} & \textbf{17.02} & \textbf{9.92} & \textbf{13.58} \\
        E3M3 & 6.89 & 9.78 & 8.43 & 11.27 & 14.01 & 17.26 & 10.02 & 13.71 \\
        E2M4 & 12.66 & 20.41 & 10.83 & 15.28 & 14.63 & 10.16 & 10.43 & 14.26 \\
        E3M2 & 8.18 & 11.74 & 9.50 & 12.54 & 14.47 & 17.88 & 10.29 & 14.09 \\
        E2M3 & 15.76 & 26.49 & 13.13 & 17.51 & 15.01 & 18.47 & 10.48 & 14.33 
      \\
      \bottomrule
    \end{tabular}
  \label{tab:a_scale_format}
\end{table}

\subsection{Redundancy in NVFP4 Block Element Encoding} \label{sec:nvfp4_element_redun}
Based on Eq.~\ref{eq:fp4_value}, the FP4 format can represent the set of quantization values $\pm\{0, 0.5, 1, 1.5, 2, 3, 4, 6\}$, including both positive zero and negative zero. This redundancy leads to underutilization of the FP4 representational space. Our proposed \workname{} format aims to customize the FP4 format for each block by repurposing the redundant zero as a special value. To determine the per-block special value, \workname{} first defines a set of allowed special values, and for each block, it replaces the redundant zero encoding of FP4 with one special value, such that the block error is minimized. This optimization process can be formulated as:
\begin{equation}
    v_{i} = \argmin_{v \,\in\, V} \norm{
      \left\lfloor X^{\text{Scaled}}_{i} ,\, Q^{\text{FP4}} \cup \{v\} \right\rceil - X^{\text{Scaled}}_{i}
    }^2_2
    \label{eq:sv_optiimize}
\end{equation} 
\begin{equation} 
    \bar{X}^{\text{RaZeR}}_{i} = \left\lfloor
        X^{\text{Scaled}}_{i} \,,\, Q^{\text{FP4}} \cup \{v_i\}
    \right\rceil
    \label{eq:razer_quant}
\end{equation} 
where $V$ is the set of allowed special values, and $v_i$ is the optimal special value for block $i$, $X^{\text{Scaled}}_{i}$ is the 16-element block after tensor-wise and block-wise scaling, and $\bar{X}^{\text{RaZeR}}_{i}$ is the quantized \workname{} block. In order to apply Eq.~\ref{eq:sv_optiimize}, we need to address two key questions: (1) How many special values should be allowed for selection? (2) What values should be included in the set of allowed special values?

To answer the first question, it is important to understand the trade-off between the number of special values and the encoding cost required to select among them. With $N$ different possible special values, selecting one special value per block incurs an encoding overhead of $\ceil[\big]{\text{log}N}$ bits. As described in Sec.~\ref{sec:nvfp4_scale_redun}, the block scales for weight and activation have a redundancy of 2 bits and 1 bit, respectively. Utilizing these redundant bits in the block scales as encoding for special value selection, \workname{} can support 4 special values for weights and 2 for activations, while preserving the same memory footprint as NVFP4. 
For deployment, the per-block special value for LLM weights can be precomputed offline. For activations, however, applying Eq.~\ref{eq:sv_optiimize} requires dynamically quantizing each block twice---once for each of the two special values---and comparing their mean squared quantization errors. Fortunately, the overhead of online double quantization is minimal and amounts to less than $2\%$ time of the NVFP4 quantizer, as demonstrated by~\citet{4over6}.

\begin{figure}[t]
  \begin{center}
    \centerline{\includegraphics[width=\columnwidth]{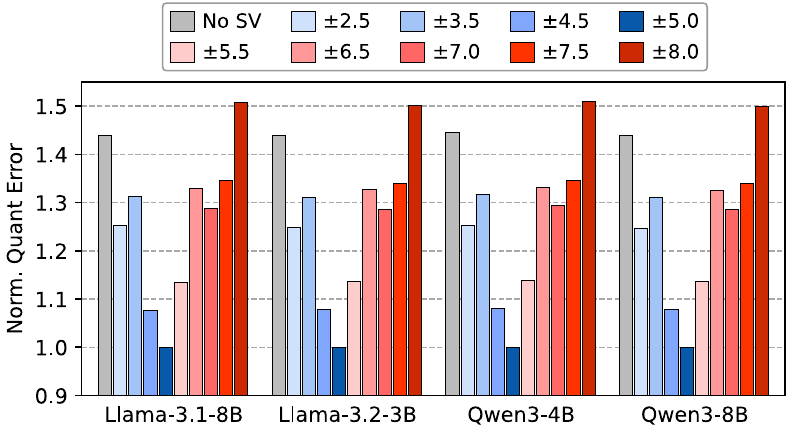}}
    \caption{
      Normalized weight quantization error ($\downarrow$) with respect to adding different special value (SV) pairs to NVFP4. The gray bar with no special value represents the baseline NVFP4.
    }
    \label{fig:fp4_sv_error}
  \end{center}
  \vspace{-25pt}
\end{figure}

To answer the second question, it is essential to consider the hardware implication of special values during runtime computation. If a special value can represent arbitrary numbers in FP16 or BF16, the corresponding MAC would need to support high-precision arithmetic. However, this undermines a key benefit of NVFP4---the area and energy efficiency brought by the low-precision arithmetic. Aiming to preserve this benefit, we restrict the special value to be a multiple of 0.5, maintaining the same numerical granularity as FP4 and remaining compatible with low-precision hardware. Additionally, we group the number of allowed special values into pairs, where every pair has the same absolute value but opposite signs. This can further reduce the hardware cost (Sec.~\ref{sec:nvfp4_mac}) to support \workname{}. Given these design considerations, the search space of allowed special values is significantly reduced, allowing for a sweep on the possible sets of  special values to find the configuration with the lowest quantization error. For weights, this can be done in an offline manner. For activations, we follow prior works~\cite{smoothquant, awq} and use a calibration set from the Pile dataset~\cite{pile} to determine the two allowed special values.

Fig.~\ref{fig:fp4_sv_error} shows the normalized weight quantization error of four LLMs, when adding one pair of allowed special values to NVFP4. We observe that the quantization error generally follows a parabolic trend as the absolute value increases, with the lowest error occurring at $\pm\,5$ across all models. Interestingly, we find that the same trend holds for activation quantization. This is because $\pm\,5$ lies at the midpoint between FP4's two largest representable values (i.e., $\pm\,4$ and $\pm\,6$), and adding it effectively bridges the gap to enable more accurate quantization of near-maximal values. For the weights, we can choose from 4 special values, and hence we measure the quantization error of adding another special value pair on top of $\pm\,5$. The optimal set of weight special values for each model is presented in Appendix~\ref{apdx:razer_detail}.

\subsection{GPU Kernel Implementation for \workname{}}
We demonstrate the feasibility of deploying \workname{} on existing hardware through a weight-only quantized GEMM kernel optimized for Blackwell GPUs~\cite{blackwell-gpu}. Inspired by the Marlin kernel~\cite{marlin}, we adopt a similar execution mechanism to develop an FP4$\times$FP16 kernel with support for redundant zero remapping, targeting high performance across both low-batch and high-batch regimes. We additionally implement a CUDA-core variant to further reduce latency for GEMV workloads beyond the tensor-core implementation.
We use a block size of 128 and shuffle the quantized FP4 weights offline so that the load process can load continuous memory data and exactly match each thread’s tensor data-fragment requirements. Every weight block is paired with a single FP16 scaling factor for dequantization. Similar to RaZeR for NVFP4, we exploit redundancy in the FP16 block scale by using its sign bit and most significant exponent bit to store the 2-bit metadata for selecting special values.

For the tensor-core-based kernel, we employ a work-partitioning strategy similar to Marlin~\cite{marlin}: each SM executes one thread block, and each thread block processes a stripe of weights with approximately equal length, potentially spanning multiple $N$-dimensions. A final reduction stage aggregates partial results across thread blocks. This ensures that each block has a similar execution time and mitigates tail effects. 
However, we observed that modern GPUs~\cite{blackwell-gpu} contain a very large number of streaming multi-processors (SMs) (e.g., 188 SMs on an RTX Pro 6000). For small weight matrices, this leads to excessive partitioning of the workload, making the sequential reduction stage a non-negligible overhead. To address this issue, we introduce an additional auto-tuning strategy that limits the number of participating SMs and thread blocks, thereby partitioning the weight matrix into larger chunks. The strategy performs offline profiling for each matrix size and, at inference time, selects the optimal number of SMs to launch for each batch size. Although this approach reduces theoretical peak computational throughput, it reduces reduction overhead and lowers GEMM latency for small batch sizes and small weight matrices, achieving up to 9.87\% throughput improvement on multiple model inference benchmarks. Appendix~\ref{apdx:E} gives a more detailed explanation of the mechanism and profiling data on this optimization.

In the Appendix~\ref{apdx:w4a4_tensorcore_realization}, we further investigate the implementation of 4-bit weight-activation \workname{} GEMM on Blackwell GPU tensor cores operating in NVFP4 mode with a finer block size of 16. Because GPU tensor cores lack native support for this remap mechanism, we propose a novel two-pass method to achieve equivalent computation.

\begin{figure}[t]
  \begin{center}
    \centerline{\includegraphics[width=\columnwidth]{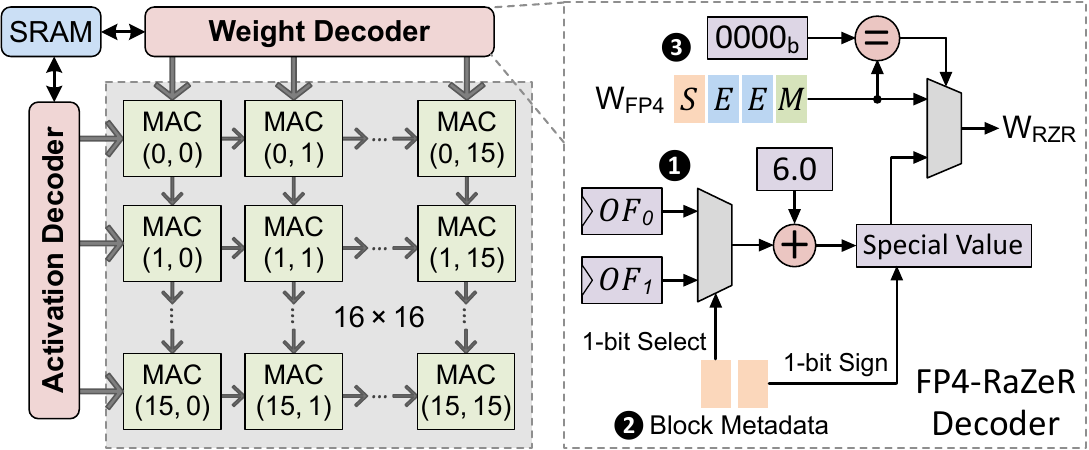}}
    \caption{
      Overview of the \workname{} tensor core. The weight decoder converts an FP4 value to the actual \workname{} value. The activation decoder is similar to the weight decoder, but has only one offset (OF) register and does not require the 1-bit select signal. 
    }
    \label{fig:razer_hardware}
  \end{center}
  \vspace{-20pt}
\end{figure}

\begin{table*} [t]
  \centering
  \setlength{\tabcolsep}{2pt}
  \renewcommand{\arraystretch}{1.25}
  \footnotesize
  \caption{Perplexity ($\downarrow$) under different 4-bit weight-only and weight-activation quantization methods. KV-cache remains in FP16.}
  \vspace{-5pt}
    \begin{tabular}{ c | c || c  c : c  c : c  c : c  c || c  c : c  c : c  c : c  c || c }
      \Xhline{0.2ex}
        \multirow[b]{3}{*}{\shortstack{\textbf{\#Bits} \vspace{4.5pt} \\ \text{\scriptsize W-A}}} & \multirow{3}{*}{\textbf{Method}} & \multicolumn{8}{c||}{\textbf{Llama}} & \multicolumn{8}{c||}{\textbf{Qwen}} & \multirow{3}{*}{\textbf{Avg.}} \\
        & & \multicolumn{2}{c:}{2-7B} & \multicolumn{2}{c:}{2-13B} & \multicolumn{2}{c:}{3.1-8B} & \multicolumn{2}{c||}{3.2-3B} & \multicolumn{2}{c:}{3-4B} & \multicolumn{2}{c:}{3-8B} & \multicolumn{2}{c:}{3-14B} & \multicolumn{2}{c||}{3-32B} & \\
        & & Wiki & C4 & Wiki & C4 & Wiki & C4 & Wiki & C4 & Wiki & C4 & Wiki & C4 & Wiki & C4 & Wiki & C4 & \\
      \Xhline{0.2ex}\noalign{\vskip 1pt}
        16-16 & FP16 & 5.47 & 6.97 & 4.88 & 6.47 & 6.24 & 8.96 & 7.82 & 10.44 & 13.66 & 16.65 & 9.73 & 13.30 & 8.65 & 12.02 & 7.61 & 10.78 & 9.35 \\
      \hdashline\noalign{\vskip 1pt}
        \multirow{7}{*}{4-16} & MXFP4 & 5.91 & 7.40 & 5.17 & 6.71 & 7.04 & 10.05 & 8.65 & 11.59 & 15.01 & 18.10 & 10.53 & 14.22 & 9.39 & 12.70 & 7.93 & 11.23 & 10.10 \\
        & NVFP4 & 5.63 & 7.15 & 4.98 & 6.56 & 6.63 & 9.48 & 8.24 & 11.10 & 13.83 & 16.85 & 9.92 & 13.55 & 8.80 & 12.23 & 7.87 & 10.98 & 9.61 \\
        & GPTQ & 5.64 & 7.13 & 4.97 & 6.56 & 6.59 & 9.48 & 8.29 & 11.27 & 13.96 & 16.94 & 10.20 & 13.68 & 8.76 & 12.21 & 7.75 & 10.95 & 9.65 \\
        & AWQ & 5.60 & 7.11 & 4.97 & 6.55 & 6.62 & 9.45 & 8.23 & 11.04 & 14.57 & 17.59 & 9.96 & 13.60 & 8.87 & 12.30 & 7.78 & 10.97 & 9.70 \\
        & SqueezeLLM & 5.61 & 7.11 & 4.99 & 6.56 & 6.64 & 9.49 & 8.24 & 11.05 & 14.16 & 17.27 & 10.07 & 13.71 & 8.98 & 12.34 & 7.83 & 11.02 & 9.69 \\
        & 4over6 & 5.61 & 7.14 & 4.97 & 6.55 & 6.60 & 9.42 & 8.21 & 11.05 & 13.83 & 16.85 & 9.89 & 13.54 & \textbf{8.72} & 12.19 & 7.80 & 11.04 & 9.59 \\
        & \cellcolor{red!10}\textbf{\workname{}} & \cellcolor{red!10}\textbf{5.57} & \cellcolor{red!10}\textbf{7.09} & \cellcolor{red!10}\textbf{4.95} & \cellcolor{red!10}\textbf{6.53} & \cellcolor{red!10}\textbf{6.50} & \cellcolor{red!10}\textbf{9.29} & \cellcolor{red!10}\textbf{8.10} & \cellcolor{red!10}\textbf{10.86} & \cellcolor{red!10}\textbf{13.83} & \cellcolor{red!10}\textbf{16.85} & \cellcolor{red!10}\textbf{9.75} & \cellcolor{red!10}\textbf{13.46} & \cellcolor{red!10}8.77 & \cellcolor{red!10}\textbf{12.14} & \cellcolor{red!10}\textbf{7.74} & \cellcolor{red!10}\textbf{10.89} & \cellcolor{red!10}\textbf{9.52} \\ 
      \hdashline\noalign{\vskip 1pt}
        \multirow{7}{*}{4-4} & MXFP4 & 7.01 & 8.67 & 5.83 & 7.49 & 8.29 & 11.65 & 10.18 & 13.49 & 17.46 & 20.24 & 11.84 & 15.77 & 10.20 & 13.62 & 8.49 & 11.87 & 11.38 \\
        & NVFP4 & 5.77 & 7.29 & 5.07 & 6.66 & 6.95 & 9.94 & 8.60 & 11.65 & 13.88 & 17.21 & 10.05 & 13.75 & 8.92 & 12.38 & 7.99 & 11.14 & 9.83 \\
        & NF4 & 5.83 & 7.37 & 5.14 & 6.70 & 7.14 & 10.21 & 8.87 & 11.93 & 14.99 & 18.07 & 10.20 & 14.06 & 9.16 & 12.63 & 8.23 & 11.62 & 10.13 \\
        & BlockDialect & 5.76 & 7.27 & 5.08 & 6.65 & 6.89 & 9.87 & 8.60 & 11.57 & 14.48 & 17.52 & 10.34 & 13.99 & 9.13 & 12.46 & 7.92 & 11.12 & 10.13 \\
        & MR-GPTQ & 5.74 & 7.25 & 5.07 & 6.65 & 6.89 & 9.92 & 8.64 & 12.18 & 14.11 & 17.27 & 10.26 & 13.85 & 8.95 & 12.38 & 7.87 & 11.09 & 9.88 \\
        & 4over6 & 5.72 & 7.25 & 5.06 & 6.63 & 6.88 & 9.83 & 8.53 & 11.53 & 13.88 & 17.21 & 10.00 & 13.73 & \textbf{8.83} & 12.33 & 7.91 & 11.14 & 9.78 \\
        & \cellcolor{red!10}\textbf{\workname{}} & \cellcolor{red!10}\textbf{5.66} & \cellcolor{red!10}\textbf{7.19} & \cellcolor{red!10}\textbf{5.01} & \cellcolor{red!10}\textbf{6.59} & \cellcolor{red!10}\textbf{6.74} & \cellcolor{red!10}\textbf{9.63} & \cellcolor{red!10}\textbf{8.37} & \cellcolor{red!10}\textbf{11.27} & \cellcolor{red!10}\textbf{13.82} & \cellcolor{red!10}\textbf{17.11} & \cellcolor{red!10}\textbf{9.83} & \cellcolor{red!10}\textbf{13.60} & \cellcolor{red!10}8.86 & \cellcolor{red!10}\textbf{12.26} & \cellcolor{red!10}\textbf{7.84} & \cellcolor{red!10}\textbf{11.02} & \cellcolor{red!10}\textbf{9.68} \\
      \Xhline{0.2ex}
    \end{tabular}
  \label{tab:ppl_exp}
\end{table*}

\begin{table*} [t]
  \centering
  \setlength{\tabcolsep}{6pt}
  \renewcommand{\arraystretch}{1.2}
  \footnotesize
  \caption{Average zero-shot accuracy ($\uparrow$) on multiple tasks under different 4-bit weight-activation quantization methods. KV-cache remains in FP16. The detailed results for each task are presented in Appendix~\ref{apdx:zeroshot_exp}.}
  \vspace{-5pt}
    \begin{tabular}{ c | c || c c c c || c c c c || c }
      \Xhline{0.2ex}
        \multirow[b]{2}{*}{\shortstack{\textbf{\#Bits} \\ \text{\scriptsize W-A}}} & \multirow{2}{*}{\textbf{Method}} & \multicolumn{4}{c||}{\textbf{Llama}} & \multicolumn{4}{c||}{\textbf{Qwen}} & \multirow{2}{*}{\textbf{Avg.}} \\
        & & {2-7B} & {2-13B} & {3.1-8B} & {3.2-3B} & {3-4B} & {3-8B} & {3-14B} & {3-32B} & \\
      \Xhline{0.2ex}\noalign{\vskip 1pt}
        16-16 & FP16 & 69.49 & 72.33 & 74.42 & 68.20 & 66.84 & 70.37 & 73.82 & 74.86 & 71.29 \\
      \hdashline\noalign{\vskip 1pt}
        \multirow{5}{*}{4-4} & MXFP4 & 65.57 & 69.39 & 67.61 & 62.41 & 61.13 & 65.80 & 70.52 & 73.45 & 66.98 \\
        & NVFP4 & 68.30 & 71.41 & 72.27 & 66.06 & 63.28 & 69.15 & 72.02 & 73.62 & 69.51 \\
        & MR-GPTQ & 68.69 & 71.59 & 72.70 & 66.18 & 63.89 & 68.95 & 72.45 & 74.11 & 69.82 \\
        & 4over6 & 68.42 & 71.34 & \textbf{73.16} & 66.00 & 63.74 & 68.71 & 72.54 & 74.23 & 69.77 \\
        & \cellcolor{red!10}\textbf{\workname{}} & \cellcolor{red!10}\textbf{68.89} & \cellcolor{red!10}\textbf{72.12} & \cellcolor{red!10}72.98 & \cellcolor{red!10}\textbf{66.65} & \cellcolor{red!10}\textbf{63.93} & \cellcolor{red!10}\textbf{69.23} & \cellcolor{red!10}\textbf{72.86} & \cellcolor{red!10}\textbf{74.49} & \cellcolor{red!10}\textbf{70.14} \\
      \Xhline{0.2ex}
    \end{tabular}
  \label{tab:zeroshot_exp}
\end{table*}

\subsection{Tensor Core Architecture for \workname{}} \label{sec:nvfp4_mac}
Since the current NVFP4 tensor cores on Blackwell GPUs do not natively support \workname{} for 4-bit GEMM, we design a custom \workname{} tensor core (Fig.~\ref{fig:razer_hardware}) to demonstrate its practicality on future GPU platforms. We assume $16 \times 16$ MAC units organized in a single-instruction-multiple-data (SIMD) way to align with the block size of NVFP4. The on-chip SRAM sends data to the weight and activation decoders, which convert a FP4 binary input to the corresponding RaZeR value before performing low-precision MAC operations. 

Taking the weight decoder as an example, recall from Sec.~\ref{sec:nvfp4_element_redun} that \workname{} supports four special values, organized into two pairs of additive inverses. To minimize decoder area overhead, we employ an offset-based encoding to store and select these values, as shown in Fig.~\ref{fig:razer_hardware}. ~\protect\circled{\small1} The weight decoder stores the four special values using two 4-bit offset registers ($OF_0$ and $OF_1$). Each register encodes a signed fixed-point value with 1 sign bit, 2 integer bits, and 1 fraction bit, representing values in $[-3.5, 3.5]$ with a step size of $0.5$. ~\protect\circled{\small2} At decode time, a 1-bit selector from the metadata chooses the target offset, which is added to $6.0$ (the maximum FP4 value) to reconstruct the magnitude. The final special value is obtained by concatenating a 1-bit sign from the metadata. For example, to produce the special value $-5.0$, an offset register stores $1010_b$ (i.e., $-1.0$); adding this offset to $6.0$ yields $5.0$, which is then combined with a negative sign bit. ~\protect\circled{\small3} During execution, the FP4 weight $W_{\text{FP4}}$ is compared against binary zero; upon a match, the decoder outputs the reconstructed \workname{} weight $W_{\text{RZR}}$.

\section{Evaluations} \label{sec:experiments}
\subsection{Settings} \label{sec:exp_setting}
\noindent
\textbf{Models.}
We conduct comprehensive evaluations of \workname{} using models from Llama and Qwen3 families with varying sizes. Specifically, we test \text{Llama-2}-\{7B, 13B\}~\cite{llama-2}, \text{Llama-3}-\{3B, 8B\}~\cite{llama-3}, and \text{Qwen3}-\{4B, 8B, 14B, 32B\}~\cite{qwen3}. We start from the original HuggingFace Transformers implementations of these models and quantize the model weights using the \workname{} procedure described in Eq.~\ref{eq:sv_optiimize}--\ref{eq:razer_quant}.

\noindent
\textbf{Datasets.}
We evaluate the perplexity on Wikitext-2~\cite{wikitext} and C4~\cite{c4} datasets using a default context length of 2K. We evaluate the accuracy on LM-Eval~\cite{lm-eval} zero-shot tasks, including PIQA~\cite{piqa}, HellaSwag~\cite{hellaswag}, Winogrande~\cite{winogrande}, ARC~\cite{arc}, and LAMBADA~\cite{lambada}. Additionally, we evaluate two difficult mathematical and logical reasoning tasks, GSM8K~\cite{gsm8k} and MMLU~\cite{mmlu} under chain-of-thought, using the instruction-tuned variant of Llama-3.1-8B and Llama-3.2-3B. 

\noindent
\textbf{Baselines.} We compare \workname{} against previous methods on weight-only and weight-activation quantization. For weight-only quantization, we compare with vanilla MXFP4~\cite{mx-format} and NVFP4 formats, GPTQ~\cite{gptq}, AWQ~\cite{awq}, and SqueezeLLM~\cite{squeeze-llm}. For weight-activation quantization, we compare with MXFP4, NVFP4, NF4~\cite{nf4}, BlockDialect~\cite{blockdialect}, MR-GPTQ~\cite{fp_quant}, and FourOverSix~\cite{4over6}. 

\noindent
\textbf{Quantization Details.}
All baseline methods adopt block-wise quantization, except for SqueezeLLM, which only supports per-channel quantization. To promote a fair comparison, we use a block size of 32 with FP16 block scale for GPTQ, AWQ, NF4, and Atom, and a block size of 16 with 8-bit block scale for MXFP4, NVFP4, BlockDialect, MR-GPTQ, FourOverSix, and \workname{}. These configurations ensure that all compared block-wise quantization methods have the same effective precision of 4.5 bits. For \workname{}, we use $\pm\,5$ as two allowed special values for both weights and activations, as discussed in Sec.~\ref{sec:nvfp4_element_redun}. The additional two special values for weights vary across different models and are summarized in Appendix~\ref{apdx:razer_detail}.

\subsection{Accuracy Results} \label{sec:exp_accuracy}
\noindent
\textbf{Perplexity.} Table~\ref{tab:ppl_exp} summarizes the perplexity results for weight-only and weight-activation quantization across different methods. \workname{} consistently outperforms prior approaches in both quantization settings, incurring only $0.17$ and $0.33$ perplexity loss in average, respectively, relative to the FP16 baseline. Compared to two state-of-the-art methods, NVFP4 and FourOverSix, \workname{} reduces the average perplexity loss by $34.6\%\,$/$\,31.2\%$ and $29.2\%\,$/$\,23.3\%$, respectively, under weight-only$\,$/$\,$weight-activation quantization. These findings demonstrate that \workname{} can effectively adapt to a multitude of quantization configurations. Additional comparisons on joint quantization of weights, activations, and KV-cache are provided in Appendix~\ref{apdx:wakv_results}.

\noindent
\textbf{Zero-shot Accuracy.} Table~\ref{tab:zeroshot_exp} summarizes the average zero-shot task accuracy for weight-activation quantization across various methods. The detailed results for each task can be found in Appendix~\ref{apdx:zeroshot_exp}. \workname{} generally outperforms prior approaches, surpassing the baseline NVFP4 format by $0.63\%$ better accuracy averaged over all models. We note that these zero-shot tasks are generally simple, consisting of primarily multiple-choice questions. As a result, most baseline methods already achieve strong performance, while \workname{} is able to further improve the quantized model accuracy through the novel redundant zero remapping technique.

\begin{table} [t]
  \centering
  \setlength{\tabcolsep}{3pt}
  \renewcommand{\arraystretch}{1.25}
  \footnotesize
  \caption{Reasoning task accuracy ($\uparrow$) under different 4-bit weight-only and weight-activation quantization methods.}
  \vspace{-3pt}
    \begin{tabular}{ c | c || c  c : c  c || c }
      \Xhline{0.2ex}
        \multirow[b]{2}{*}{\shortstack{\textbf{\#Bits} \\ \text{\scriptsize W-A}}} & \multirow{2}{*}{\textbf{Method}} & \multicolumn{2}{c:}{\textbf{GSM8K}} & \multicolumn{2}{c||}{\textbf{MMLU}} & \multirow{2}{*}{\textbf{Avg.}} \\
        & & {3.1-8B} & {3.2-3B} & {3.1-8B} & {3.2-3B} & \\
      \Xhline{0.2ex}\noalign{\vskip 1pt}
        16-16 & FP16 & 85.52 & 78.09 & 72.65 & 65.25 & 75.37 \\
      \hdashline\noalign{\vskip 1pt}
        \multirow{5}{*}{4-16} & MXFP4 & 78.16 & 72.86 & 68.04 & 61.10 & 70.04 \\
        & NVFP4 & 82.11 & 75.06 & 70.46 & 62.04 & 72.42 \\
        & GPTQ & 83.24 & 73.31 & 70.83 & 62.57 & 72.49 \\
        & AWQ & 78.49 & 75.74 & 70.18 & 62.62 & 71.76 \\
        & \cellcolor{red!10}\textbf{\workname{}} & \cellcolor{red!10}\textbf{84.01} & \cellcolor{red!10}\textbf{77.26} & \cellcolor{red!10}\textbf{71.34} & \cellcolor{red!10}\textbf{63.10} & \cellcolor{red!10}\textbf{73.92} \\ 
      \hdashline\noalign{\vskip 1pt}
        \multirow{5}{*}{4-4} & MXFP4 & 61.49 & 57.32 & 57.97 & 55.68 & 58.11 \\
        & NVFP4 & 77.71 & 71.95 & 68.44 & 61.00 & 69.78 \\
        & MR-GPTQ & 78.43 & 71.49 & 68.85 & 61.29 & 69.99 \\
        & 4over6 & 78.17 & 72.48 & 68.99 & 61.44 & 70.27 \\
        & \cellcolor{red!10}\textbf{\workname{}} & \cellcolor{red!10}\textbf{82.64} & \cellcolor{red!10}\textbf{75.74} & \cellcolor{red!10}\textbf{69.88} & \cellcolor{red!10}\textbf{62.11} & \cellcolor{red!10}\textbf{72.59} \\
      \Xhline{0.2ex}
    \end{tabular}
  \label{tab:reasoning_exp}
  \vspace{-10pt}
\end{table}

\noindent
\textbf{Reasoning Accuracy.} Table~\ref{tab:reasoning_exp} summarizes the reasoning task accuracy for Llama-3.1-8B and Llama-3.2-3B, under different weight-only and weight-activation quantization methods. \workname{} demonstrates significant accuracy improvement over prior approaches. For instance, on Llama-3.1-8B running GSM8K, \workname{} outperforms recent NVFP4-based quantization algorithms, MR-GPTQ and FourOverSix, by a large margin of $4.21\%$ and $4.47\%$, respectively, under weight-activation quantization.

\subsection{Ablation Studies} \label{sec:algo_ablation}
\noindent
\textbf{Impact of \workname{} on Weights and Activations.} We apply \workname{} to weights and activations independently to quantify its benefits. As shown in Table~\ref{tab:ablation_wa}, \workname{} improves the accuracy of both activation-only and weight-only quantization compared to the baseline NVFP4 format. The lowest perplexity is achieved when both weights and activations adopt \workname{}. Notably, applying \workname{} only to activations achieves performance comparable to the state-of-the-art method FourOverSix, while applying it only to weights surpasses FourOverSix. These results highlight the effectiveness of \workname{} for enhanced NVFP4 quantization.

\begin{table} [t]
  \centering
  \setlength{\tabcolsep}{1.8pt}
  \renewcommand{\arraystretch}{1.25}
  \footnotesize
  \caption{Ablation study of \workname{} applied to activation-only, weight-only, and weight-activation quantization for Llama models.}
  \vspace{-3pt}
    \begin{tabular}{ c || c  c : c  c : c  c : c  c }
      \Xhline{0.2ex}
        \multirow[b]{2}{*}{\shortstack{\textbf{Format} \vspace{3.5pt} \\ \text{\footnotesize W-A}}} & \multicolumn{2}{c:}{\textbf{2-7B}} & \multicolumn{2}{c:}{\textbf{2-13B}} & \multicolumn{2}{c:}{\textbf{3.1-8B}} & \multicolumn{2}{c}{\textbf{3.2-3B}} \\
        & Wiki & C4 & Wiki & C4 & Wiki & C4 & Wiki & C4 \\
      \Xhline{0.2ex}\noalign{\vskip 1pt}
        NVFP4-NVFP4 & 5.77 & 7.29 & 5.07 & 6.66 & 6.95 & 9.94 & 8.60 & 11.65 \\
        4over6-4over6  & 5.72 & 7.25 & 5.06 & 6.63 & 6.88 & 9.83 & 8.53 & 11.53 \\
      \hdashline\noalign{\vskip 1pt}
        \workname{}-NVFP4 & 5.73 & 7.25 & 5.05 & 6.63 & 6.87 & 9.82 & 8.53 & 11.52 \\
        NVFP4-\workname{} & 5.69 & 7.22 & 5.05 & 6.62 & 6.80 & 9.73 & 8.46 & 11.38 \\
        \rowcolor{red!10}
        \textbf{\workname{}-\workname{}} & \textbf{5.66} & \textbf{7.19} & \textbf{5.01} & \textbf{6.59} & \textbf{6.74} & \textbf{9.63} & \textbf{8.37} & \textbf{11.27} \\
      \Xhline{0.2ex}
    \end{tabular}
  \label{tab:ablation_wa}
\end{table}

\begin{table} [t]
  \centering
  \setlength{\tabcolsep}{2.1pt}
  \renewcommand{\arraystretch}{1.25}
  \footnotesize
  \caption{Impact of block size on perplexity ($\downarrow$), under different 4-bit weight-activation quantization methods for Llama.}
  \vspace{-3pt}
    \begin{tabular}{ c c || c  c : c  c : c  c : c  c }
      \Xhline{0.2ex}
        \multirow{2}{*}{\textbf{Size}} & \multirow{2}{*}{\textbf{Method}} & \multicolumn{2}{c:}{\textbf{2-7B}} & \multicolumn{2}{c:}{\textbf{2-13B}} & \multicolumn{2}{c:}{\textbf{3.1-8B}} & \multicolumn{2}{c}{\textbf{3.2-3B}} \\
        & & Wiki & C4 & Wiki & C4 & Wiki & C4 & Wiki & C4 \\
      \Xhline{0.2ex}\noalign{\vskip 1pt}
        \multirow{3}{*}{16} & NVFP4 & 5.77 & 7.29 & 5.07 & 6.66 & 6.95 & 9.94 & 8.60 & 11.65 \\
        & 4over6 & 5.72 & 7.25 & 5.06 & 6.63 & 6.88 & 9.83 & 8.53 & 11.53 \\
        & \cellcolor{red!10}\textbf{\workname{}} & \cellcolor{red!10}\textbf{5.66} & \cellcolor{red!10}\textbf{7.19} & \cellcolor{red!10}\textbf{5.01} & \cellcolor{red!10}\textbf{6.59} & \cellcolor{red!10}\textbf{6.74} & \cellcolor{red!10}\textbf{9.63} & \cellcolor{red!10}\textbf{8.37} & \cellcolor{red!10}\textbf{11.27} \\
      \hdashline\noalign{\vskip 1pt}
        \multirow{3}{*}{32} & NVFP4 & 5.83 & 7.37 & 5.12 & 6.71 & 7.10 & 10.17 & 8.82 & 11.93 \\
        & 4over6 & 5.81 & 7.35 & 5.11 & 6.69 & 7.07 & 10.11 & 8.77 & 11.84 \\
        & \cellcolor{red!10}\textbf{\workname{}} & \cellcolor{red!10}\textbf{5.74} & \cellcolor{red!10}\textbf{7.27} & \cellcolor{red!10}\textbf{5.06} & \cellcolor{red!10}\textbf{6.65} & \cellcolor{red!10}\textbf{6.89} & \cellcolor{red!10}\textbf{9.88} & \cellcolor{red!10}\textbf{8.56} & \cellcolor{red!10}\textbf{11.54} \\
      \hdashline\noalign{\vskip 1pt}
        \multirow{3}{*}{64} & NVFP4 & 5.91 & 7.47 & 5.20 & 6.78 & 7.32 & 10.45 & 9.08 & 12.24 \\
        & 4over6 & 5.90 & 7.46 & 5.20 & 6.77 & 7.30 & 10.40 & 9.06 & 12.21 \\
        & \cellcolor{red!10}\textbf{\workname{}} & \cellcolor{red!10}\textbf{5.82} & \cellcolor{red!10}\textbf{7.37} & \cellcolor{red!10}\textbf{5.13} & \cellcolor{red!10}\textbf{6.71} & \cellcolor{red!10}\textbf{7.09} & \cellcolor{red!10}\textbf{10.15} & \cellcolor{red!10}\textbf{8.80} & \cellcolor{red!10}\textbf{11.90} \\
      \hdashline\noalign{\vskip 1pt}
        \multirow{3}{*}{128} & NVFP4 & 6.05 & 7.62 & 5.31 & 6.87 & 7.61 & 10.76 & 9.48 & 12.73 \\
        & 4over6 & 6.06 & 7.62 & 5.29 & 6.86 & 7.62 & 10.78 & 9.51 & 12.72 \\
        & \cellcolor{red!10}\textbf{\workname{}} & \cellcolor{red!10}\textbf{5.96} & \cellcolor{red!10}\textbf{7.52} & \cellcolor{red!10}\textbf{5.21} & \cellcolor{red!10}\textbf{6.80} & \cellcolor{red!10}\textbf{7.39} & \cellcolor{red!10}\textbf{10.50} & \cellcolor{red!10}\textbf{9.14} & \cellcolor{red!10}\textbf{12.35} \\
      \Xhline{0.2ex}
    \end{tabular}
  \label{tab:ablation_bs}
\end{table}

\noindent
\textbf{Impact of Block Size.} While the original NVFP4 format has a fixed block size of 16 and an effective bit-width of 4.5, one can flexibly adjust the block size to reduce the memory and computational overhead of block scaling. We explore the impact of block size on NVFP4, FourOverSix, and \workname{}, using different Llama models. As shown in Table~\ref{tab:ablation_bs}, \workname{} consistently achieves the best perplexity across different block sizes, thanks to its additional quantization value. In particular, the advantage of FourOverSix over NVFP4 gradually diminishes as the block size increases. This is because FourOverSix allows each block to be scaled either to the full FP4 range of $-6 \thicksim 6$, or to a narrower range of $-4 \thicksim 4$. At small block sizes, scaling to $-4 \thicksim 4$ reduces the quantization error of near-maximal values, thus improving model performance. However, with larger block sizes, scaling to $-4 \thicksim 4$ discards two available FP4 values and is therefore rarely adopted by any block, resulting in performance similar to NVFP4.

\subsection{Combination with Other Quantization Approach} \label{sec:comb_sota_quant}
We combine \workname{} with AWQ~\cite{awq}, a state-of-the-art weight-only quantization scheme, to demonstrate its potential synergy with existing approaches. Table~\ref{tab:combine_awq} reports the perplexity of various Llama models under 4-bit AWQ using a block size of 128 and different weight formats. Incorporating \workname{} significantly reduces model perplexity compared to the baseline AWQ implementation using INT4 weight format. These results highlight the robustness of \workname{} under established quantization optimizations. 

\begin{table} [t]
  \centering
  \setlength{\tabcolsep}{1.8pt}
  \renewcommand{\arraystretch}{1.25}
  \footnotesize
  \caption{Llama model perplexity ($\downarrow$) of AWQ weight-only quantization combined with different formats.}
  \vspace{-3pt}
    \begin{tabular}{ c || c  c : c  c : c  c : c  c }
      \Xhline{0.2ex}
        \multirow{2}{*}{\textbf{Method}} & \multicolumn{2}{c:}{\textbf{2-7B}} & \multicolumn{2}{c:}{\textbf{2-13B}} & \multicolumn{2}{c:}{\textbf{3.1-8B}} & \multicolumn{2}{c}{\textbf{3.2-3B}} \\
        & Wiki & C4 & Wiki & C4 & Wiki & C4 & Wiki & C4 \\
      \Xhline{0.2ex}\noalign{\vskip 1pt}
        AWQ$\,$+$\,$INT4 & 5.68 & 7.19 & 5.01 & 6.60 & 6.80 & 9.72 & 8.50 & 11.38 \\
        AWQ$\,$+$\,$FP4 & 5.63 & 7.14 & 4.99 & 6.58 & 6.68 & 9.56 & 8.32 & 11.18 \\
        \rowcolor{red!10}
        \textbf{AWQ$\,$+$\,$\workname{}} & \textbf{5.60} & \textbf{7.10} & \textbf{4.98} & \textbf{6.56} & \textbf{6.61} & \textbf{9.41} & \textbf{8.20} & \textbf{11.01} \\
      \Xhline{0.2ex}
    \end{tabular}
  \label{tab:combine_awq}
  \vspace{-10pt}
\end{table}

\subsection{GPU Kernel Performance} \label{sec:exp_kernel}
We evaluate the weight-only quantized \workname{} kernel across multiple Blackwell GPU platforms, including the datacenter-class NVIDIA RTX Pro 6000 Server Edition, the desktop edge device DGX Spark, and the RTX~5090. We compare against representative high-performance weight-only 4-bit inference kernels, including AWQ~\cite{awq}, Any-Precision-LLM~\cite{anyprecision-llm}, SqueezeLLM~\cite{squeeze-llm}, and Marlin~\cite{marlin}. These methods span both uniformly scaled INT4 quantization (e.g., AWQ, Marlin) and lookup-table--based 4-bit quantization (e.g., Any-Precision-LLM, SqueezeLLM). We also evaluate a modified Marlin variant that replaces INT4 weights with FP4 quantization in order to quantify the overhead of the redundant zero remap operation. All 4-bit kernel implementations are integrated into the \texttt{gpt-fast}~\cite{gpt-fast} framework to measure end-to-end autoregressive decoding throughput over 200 generated tokens.

\begin{figure}[t]
  \centering
  \includegraphics[width=\columnwidth]{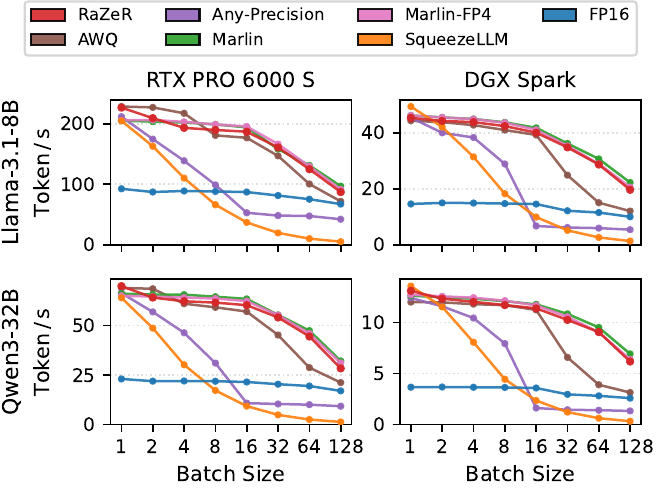}
  \caption{LLM end-to-end decode speed measurements on two representative devices. Token generation speed is reported across batch sizes.}
  \label{fig:batch_curve_small}
\end{figure}

\Cref{fig:batch_curve_small} reports the performance results on RTX Pro 6000 and DGX Spark for two representative models. Across different platforms, the \workname{} kernel achieves near-best-in-class single-batch decoding performance. At moderate to large batch sizes, its throughput is only slightly below Marlin and Marlin-FP4, while substantially outperforming AWQ and other weight-only quantized inference kernels.
\comment{
Given our \workname{} achieves much better model accuracy than plain INT4 and FP4 that used in Marlin, our kernel is at the Pareto optimal of these kernels.
}
Additional GPU evaluation results are provided in Appendix~\ref{apdx:D}, including both results on RTX 5090 and a larger diversity of models, as well as isolated kernel microbenchmarks.

\subsection{Hardware Cost Analysis} \label{sec:exp_asic}
To examine the hardware cost of adopting \workname{} on future accelerator platforms, we model the proposed custom tensor core (Fig.~\ref{fig:razer_hardware}) using SystemVerilog, which is synthesized by Synopsys Design Compiler under TSMC 28nm technology to estimate area and power. As shown in Table~\ref{tab:area_power}, the modified \workname{} tensor core incurs $3.7\%\,$/$\,13.5\%$ area$\,$/$\,$power overhead on top of the baseline NVFP4 tensor core. Given that MAC units typically occupy less than $10\%$ of the entire design within modern DNN accelerators~\cite{tpuv4}, the relative chip area$\,$/$\,$power overhead of supporting \workname{} is merely $0.37\%\,$/$\,1.35\%$.

\begin{table} [t]
  \centering
  \footnotesize
  \setlength{\tabcolsep}{2.5pt}
  \renewcommand{\arraystretch}{1.25}
  \caption{Area and power of the baseline NVFP4 tensor core and the proposed \workname{} tensor core.}
  \vspace{-3pt}
    \begin{tabular}{ c || c c c | c c c }
        \Xhline{0.3ex}
          \multirow{2}{*}{} & \multicolumn{3}{c|}{Area ($\mu$m$^2$)} & \multicolumn{3}{c}{Power (mW)} 
          \\
          & Array & Decoder & Total & Array & Decoder & Total 
          \\
        \hline
          NVFP4 & $2.315\,\text{E}5$ & $391$ & $2.319\,\text{E}5$ & 40.8 & 0.6 & 41.4 \\
          \workname{} & $2.393\,\text{E}5$ & $1201$ & $2.405\,\text{E}5$ & 45.7 & 1.3 & 47.0 \\
        \Xhline{0.3ex}
    \end{tabular}
    \vspace{-3pt}
  \label{tab:area_power}
\end{table}

\section{Conclusion} \label{sec:conclusion}

We introduce \workname{}, an enhanced NVFP4 format that exploits redundancy in the block scaling factor to repurpose the redundant FP4 zero as an optimized special value. This approach improves quantization accuracy without incurring additional memory overhead. Complementing this, we implement efficient GPU kernel and MAC unit to demonstrate the practicality of adopting \workname{} in current and future hardware platforms. Our evaluations show that \workname{} outperforms existing 4-bit methods across both weight-only and weight-activation quantization, while incurring only $0.37\%$ chip area overhead. Compared to the state-of-the-art 4-bit quantization method FourOverSix, \workname{} reduces the average perplexity loss by $29.2\%$ and $23.3\%$ under weight-only and weight-activation quantization, respectively. We hope this work can motivate future research on designing efficient low-precision numerics for LLM quantization.

\section*{Impact Statement}
This paper advances the industry-adopted NVFP4 format for LLM quantization, focusing on improving model performance while reducing memory footprint and computation energy. As a societal consequence, our work could help reduce the operational cost and energy consumption of deploying LLMs, potentially making AI systems more accessible to end users.


\bibliography{ref}
\bibliographystyle{icml2026}

\newpage
\appendix
\onecolumn

\section*{Appendix Overview}
\begin{itemize}
    \item Section~\ref{apdx:A}: Additional Ablation Study on Block Scale Format. 
    \item Section~\ref{apdx:B}: Additional Experimental Details.
    \item Section~\ref{apdx:C}: Additional Experimental Results.
    \item Section~\ref{apdx:D}: Additional Kernel Evaluation.
    \item Section~\ref{apdx:E}: Additional Experiments on Auto-tuned Kernel Performance.
\end{itemize}

\section{Additional Ablation Study on Block Scale Format} \label{apdx:A}
Table~\ref{tab:w_scale_apdx} and Table~\ref{tab:a_scale_apdx} summarize the full perplexity results when applying different block scale formats to NVFP4 weight-only and activation-only quantization, respectively. Weights are generally more tolerant to a scale format with reduced bit width. Activations are more sensitive to reduced exponent bits due to its large dynamic range. 

\begin{table} [ht]
  \centering
  \setlength{\tabcolsep}{2.5pt}
  \renewcommand{\arraystretch}{1.15}
  \footnotesize
  \caption{Wikitext-2 and C4 perplexity ($\downarrow$) using different block scale formats for weight-only NVFP4 quantization. Continuation of Table~\ref{tab:w_scale_format}. We highlight cases where the average perplexity is the same or better than that of baseline E4M3 scale.}
  \vspace{-5pt}
    \begin{tabular}{ c| c || c  c : c  c : c  c : c  c || c  c : c  c : c  c : c  c || c }
      \Xhline{0.2ex}
        \multirow{3}{*}{\textbf{\#Bits}} & \multirow{3}{*}{\textbf{Format}} & \multicolumn{8}{c||}{\textbf{Llama}} & \multicolumn{8}{c||}{\textbf{Qwen}} & \multirow{3}{*}{\textbf{Avg.}} \\
        & & \multicolumn{2}{c:}{2-7B} & \multicolumn{2}{c:}{2-13B} & \multicolumn{2}{c:}{3.1-8B} & \multicolumn{2}{c||}{3.2-3B} & \multicolumn{2}{c:}{3-4B} & \multicolumn{2}{c:}{3-8B} & \multicolumn{2}{c:}{3-14B} & \multicolumn{2}{c||}{3-32B} \\
        & & Wiki & C4 & Wiki & C4 & Wiki & C4 & Wiki & C4 & Wiki & C4 & Wiki & C4 & Wiki & C4 & Wiki & C4 \\
      \Xhline{0.2ex}
        \multirow{3}{*}{8} & E5M3 & 5.63 & 7.15 & 4.98 & 6.56 & 6.63 & 9.48 & 8.24 & 11.10 & 13.63 & 16.85 & 9.92 & 13.55 & 8.80 & 12.23 & 7.87 & 10.99 & \textbf{9.60} \\
        & E4M4 & 5.63 & 7.15 & 4.98 & 6.56 & 6.63 & 9.47 & 8.26 & 11.11 & 13.82 & 16.81 & 10.09 & 13.58 & 8.78 & 12.22 & 7.84 & 10.98 & 9.62 \\
        & E3M5 & 5.63 & 7.14 & 4.98 & 6.56 & 6.63 & 9.47 & 8.26 & 11.11 & 13.67 & 16.84 & 9.97 & 13.53 & 8.80 & 12.21 & 7.85 & 10.98 & \textbf{9.60} \\
      \hdashline\noalign{\vskip 1pt}
        \multirow{3}{*}{7} & E5M2 & 5.65 & 7.15 & 4.99 & 6.58 & 6.67 & 9.53 & 8.29 & 11.17 & 13.96 & 16.89 & 9.99 & 13.59 & 8.82 & 12.29 & 7.85 & 10.97 & 9.65 \\
        & E4M3 & 5.63 & 7.15 & 4.98 & 6.56 & 6.63 & 9.48 & 8.24 & 11.10 & 13.63 & 16.85 & 9.92 & 13.55 & 8.80 & 12.23 & 7.87 & 10.99 & \textbf{9.60} \\
        & E3M4 & 5.63 & 7.15 & 4.98 & 6.56 & 6.63 & 9.47 & 8.26 & 11.11 & 13.63 & 16.81 & 9.92 & 13.60 & 8.78 & 12.22 & 7.84 & 10.97 & \textbf{9.60} \\
      \hdashline\noalign{\vskip 1pt}
        \multirow{3}{*}{6} & E4M2 & 5.65 & 7.15 & 4.99 & 6.58 & 6.67 & 9.53 & 8.29 & 11.17 & 13.96 & 16.89 & 9.99 & 13.59 & 8.82 & 12.29 & 7.85 & 10.97 & 9.65 \\
        & E3M3 & 5.63 & 7.15 & 4.97 & 6.56 & 6.63 & 9.48 & 8.24 & 11.10 & 13.63 & 16.85 & 9.92 & 13.55 & 8.80 & 12.24 & 7.87 & 10.99 & \textbf{9.60} \\
        & E2M4 & 5.64 & 7.16 & 4.99 & 6.57 & 6.64 & 9.5 & 8.25 & 11.11 & 14.13 & 16.97 & 10.11 & 13.65 & 8.83 & 12.29 & 7.85 & 10.97 & 9.67 \\
      \hdashline\noalign{\vskip 1pt}
        \multirow{2}{*}{5} & E3M2 & 5.65 & 7.15 & 4.99 & 6.58 & 6.67 & 9.53 & 8.29 & 11.17 & 13.96 & 16.9 & 10.00 & 13.59 & 8.82 & 12.3 & 7.85 & 10.97 & 9.65 \\
        & E2M3 & 5.69 & 7.19 & 5.02 & 6.61 & 6.72 & 9.63 & 8.32 & 11.21 & 14.22 & 17.22 & 10.22 & 13.79 & 9.02 & 12.48 & 7.93 & 11.13 & 9.78 \\
      \bottomrule
    \end{tabular}
  \label{tab:w_scale_apdx}
\end{table}

\begin{table} [ht]
  \centering
  \setlength{\tabcolsep}{2.5pt}
  \renewcommand{\arraystretch}{1.15}
  \footnotesize
  \caption{Wikitext-2 and C4 perplexity ($\downarrow$) using different block scale formats for activation-only NVFP4 quantization. Continuation of Table~\ref{tab:a_scale_format}. We highlight cases where the average perplexity is the same or better than that of baseline E4M3 scale.}
  \vspace{-5pt}
    \begin{tabular}{ c| c || c  c : c  c : c  c : c  c || c  c : c  c : c  c : c  c || c }
      \Xhline{0.2ex}
        \multirow{3}{*}{\textbf{\#Bits}} & \multirow{3}{*}{\textbf{Format}} & \multicolumn{8}{c||}{\textbf{Llama}} & \multicolumn{8}{c||}{\textbf{Qwen}} & \multirow{3}{*}{\textbf{Avg.}} \\
        & & \multicolumn{2}{c:}{2-7B} & \multicolumn{2}{c:}{2-13B} & \multicolumn{2}{c:}{3.1-8B} & \multicolumn{2}{c||}{3.2-3B} & \multicolumn{2}{c:}{3-4B} & \multicolumn{2}{c:}{3-8B} & \multicolumn{2}{c:}{3-14B} & \multicolumn{2}{c||}{3-32B} \\
        & & Wiki & C4 & Wiki & C4 & Wiki & C4 & Wiki & C4 & Wiki & C4 & Wiki & C4 & Wiki & C4 & Wiki & C4 \\
      \Xhline{0.2ex}
        \multirow{3}{*}{8} & E5M3 & 5.59 & 7.09 & 4.97 & 6.55 & 6.53 & 9.35 & 8.11 & 10.9 & 13.85 & 16.93 & 9.84 & 13.5 & 8.76 & 12.16 & 7.74 & 10.93 & \textbf{9.55} \\
        & E4M4 & 5.58 & 7.08 & 4.96 & 6.55 & 6.51 & 9.33 & 8.11 & 10.87 & 13.81 & 16.9 & 9.83 & 13.47 & 8.74 & 12.15 & 7.72 & 10.92 & \textbf{9.53} \\
        & E3M5 & 5.61 & 7.11 & 4.96 & 6.55 & 6.53 & 9.36 & 8.12 & 10.9 & 13.80 & 16.89 & 9.85 & 13.48 & 8.73 & 12.15 & 7.72 & 10.93 & \textbf{9.54} \\
      \hdashline\noalign{\vskip 1pt}
        \multirow{3}{*}{7} & E5M2 & 5.62 & 7.12 & 4.99 & 6.58 & 6.58 & 9.43 & 8.18 & 10.98 & 13.97 & 17.03 & 9.91 & 13.57 & 8.81 & 12.20 & 7.77 & 10.97 & 9.61 \\
        & E4M3 & 5.59 & 7.09 & 4.97 & 6.55 & 6.53 & 9.35 & 8.12 & 10.89 & 13.81 & 16.93 & 9.84 & 13.49 & 8.77 & 12.16 & 7.74 & 10.93 & \textbf{9.55} \\
        & E3M4 & 5.67 & 7.18 & 4.97 & 6.55 & 6.61 & 9.43 & 8.18 & 10.97 & 13.89 & 16.97 & 9.89 & 13.53 & 8.74 & 12.16 & 7.73 & 10.93 & 9.59 \\
      \hdashline\noalign{\vskip 1pt}
        \multirow{3}{*}{6} & E4M2 & 5.62 & 7.12 & 4.98 & 6.57 & 6.59 & 9.43 & 8.18 & 10.98 & 13.94 & 17.02 & 9.92 & 13.58 & 8.80 & 12.21 & 7.77 & 10.97 & 9.61 \\
        & E3M3 & 5.84 & 7.32 & 5.0 & 6.58 & 6.89 & 9.78 & 8.43 & 11.27 & 14.01 & 17.26 & 10.02 & 13.71 & 8.77 & 12.23 & 7.71 & 10.97 & 9.74 \\
        & E2M4 & 17.47 & 8.43 & 5.58 & 7.37 & 12.66 & 20.41 & 11.83 & 15.28 & 14.63 & 18.16 & 10.43 & 14.26 & 8.99 & 12.52 & 7.87 & 11.33 & 12.33 \\
      \hdashline\noalign{\vskip 1pt}
        \multirow{2}{*}{5} & E3M2 & 6.75 & 7.68 & 5.17 & 6.75 & 8.18 & 11.74 & 9.50 & 12.54 & 14.47 & 17.88 & 10.29 & 14.09 & 8.91 & 12.33 & 7.77 & 11.08 & 10.32 \\
        & E2M3 & 15.93 & 9.59 & 6.36 & 8.39 & 15.77 & 26.49 & 13.13 & 17.51 & 15.01 & 18.47 & 10.48 & 14.33 & 9.22 & 12.81 & 8.10 & 11.50 & 13.32 \\
      \bottomrule
    \end{tabular}
  \label{tab:a_scale_apdx}
\end{table}
\newpage

\section{Additional Experimental Details} \label{apdx:B}

\subsection{Details for Baseline Methods}
We provide additional details about the baseline quantization methods used in our evaluation. 
\vspace{-5pt}
\begin{itemize}
    \item MXFP4: A variant of the microscaling (MX) format, which is standardized by the~\citet{mx-format}. A concrete MX-compliant format defines three items: the data format, the block size, and the block scale format. By default, the MX block size is 32 with a block scale format of E8M0. The data format can be INT8, FP8-E5M2, FP8-E4M3, FP6-E3M2, FP6-E2M3, FP4-E2M1. 
    \item NVFP4: A custom FP4 format introduced since Blackwell GPU~\cite{nvfp4}. It applies a fine-grained FP8-E4M3 scaling factor to each 16-value block, while also leveraging a second-level FP32 scaling factor applied per tensor. 
    \item NF4: The 4-bit NormalFloat format introduced in QLoRA~\cite{nf4}. It leverages the statistical quantile function to derive the 16 quantization values that are information theoretically optimal for normally distributed tensors. The 16 quantization values are stored in a high-precision BF16 lookup table. During computation, every quantized 4-bit element serves as an index to extract the corresponding BF16 value from the lookup table.
    \item BlockDialect~\cite{blockdialect}: A 4-bit block-wise mixed format that enhances MXFP4. It assigns a per-block optimal number format from a formatbook to achieve better data representation. To preserve compatibility with energy-efficient low-precision arithmetic, it introduces DialectFP4, a formatbook of FP4 variants that adapt to diverse data distributions.
    \item GPTQ~\cite{gptq}: A weight-only quantization method that uses the second-order Hessian information to perform error compensation.
    \item AWQ~\cite{awq}: A weight-only quantization method that protects the salient weight channels corresponding to larger activation magnitudes. It performs adaptive scaling to scale up the salient weight channels, which can reduce their relative quantization error. In addition, it applies weight clipping to further minimize the quantization error.
    \item SqueezeLLM~\cite{squeeze-llm}: A weight-only quantization method that (1) Leverage sensitivity-based non-uniform quantization to search for the optimal bit precision assignment based on second-order information. (2) Employ dense-and-sparse decomposition that stores outlier weights in FP16 using an efficient sparse format.
    \item MR-GPTQ~\cite{fp_quant}: A weight-activation quantization method that applies Hadamard transformation to activations, while using GPTQ to reduce the quantization error of weights. 
    \item FourOverSix~\cite{4over6}: A modification to the NVFP4 quantization algorithm that evaluates two potential scale factors for each block of values, one by scaling to the full range of FP4 values $-6 \thicksim 6$, and another by scaling to a narrower range of $-4 \thicksim 4$. It then selects the optimal scaling based on the mean-squared quantization error.
\end{itemize}

\subsection{Quantization Details for \workname} \label{apdx:razer_detail}
In our evaluation, both weights and activations use $\pm\,5$ in their set of allowed special values. The weights allow two extra special values, which vary across different models as summarized in Table~\ref{tab:sv_weight}.

\begin{table*} [ht]
  \centering
  \setlength{\tabcolsep}{5pt}
  \renewcommand{\arraystretch}{1.25}
  \footnotesize
  \caption{Four special values for \workname{} weights in different models.}
  \vspace{-5pt}
    \begin{tabular}{ c c c c || c c c c }
      \Xhline{0.2ex}
        \multicolumn{4}{c||}{\textbf{Llama}} & \multicolumn{4}{c}{\textbf{Qwen}} \\
        {2-7B} & {2-13B} & {3.1-8B} & {3.2-3B} & {3-4B} & {3-8B} & {3-14B} & {3-32B} \\
      \Xhline{0.2ex}
        $\pm\,5 \,, \pm\,8$ & $\pm\,5 \,, \pm\,8$ & $\pm\,5 \,, \pm\,8$ & $\pm\,5 \,, \pm\,8$ & $\pm\,5 \,, \pm\,8$ & $\pm\,5 \,, \pm\,7$ & $\pm\,5 \,, \pm\,8$ & $\pm\,5 \,, \pm\,9$ \\
      \Xhline{0.2ex}
    \end{tabular}
  \label{tab:sv_weight}
\end{table*}
\newpage

\section{Additional Experimental Results} \label{apdx:C}
\subsection{Perplexity under Joint Quantization of Weights, Activations, and KV-cache} \label{apdx:wakv_results}
We compare the perplexity of different 4-bit quantization methods under joint quantization of weights, activations, and KV-cache. Specifically, we compare with MXFP4~\cite{mx-format}, NVFP4~\cite{nvfp4}, NF4~\cite{nf4}, Atom~\cite{atom}, DuQuant~\cite{duquant}, SpinQuant~\cite{spinquant}, OSTQuant~\cite{ostquant}, and FourOverSix~\cite{4over6}. Since some baseline methods do not support Qwen, we focus on evaluating Llama models as shown in Table~\ref{tab:ppl_wakv}. \workname{} delivers better performance across all evaluated models and datasets. 

\begin{table} [ht]
  \centering
  \setlength{\tabcolsep}{4pt}
  \renewcommand{\arraystretch}{1.2}
  \footnotesize
  \caption{Perplexity ($\downarrow$) under joint quantization of weights, activations, and KV-cache, using different 4-bit quantization methods.}
  \vspace{-5pt}
    \begin{tabular}{ c || c  c : c  c : c  c || c  c : c  c : c  c || c  c : c  c }
      \Xhline{0.2ex}
        \multirow{3}{*}{\textbf{Method}} & \multicolumn{6}{c||}{\textbf{Llama-1}} & \multicolumn{6}{c||}{\textbf{Llama-2}} & \multicolumn{4}{c}{\textbf{Llama-3}} \\
        & \multicolumn{2}{c:}{1-7B} & \multicolumn{2}{c:}{1-13B} & \multicolumn{2}{c||}{1-65B} & \multicolumn{2}{c:}{2-7B} & \multicolumn{2}{c:}{2-13B} & \multicolumn{2}{c||}{2-70B} & \multicolumn{2}{c:}{3.1-8B} & \multicolumn{2}{c}{3.2-3B} \\
        & Wiki & C4 & Wiki & C4 & Wiki & C4 & Wiki & C4 & Wiki & C4 & Wiki & C4 & Wiki & C4 & Wiki & C4 \\
      \Xhline{0.2ex}\noalign{\vskip 1pt}
        FP16 & 5.68 & 7.08 & 5.09 & 6.61 & 3.53 & 5.62 & 5.47 & 6.97 & 4.88 & 6.46 & 3.31 & 5.52 & 6.24 & 8.96 & 7.82 & 10.44 \\
      \hdashline\noalign{\vskip 1pt}
        MXFP4 & 7.91 & 9.75 & 6.39 & 8.01 & 4.80 & 6.89 & 8.33 & 11.39 & 7.06 & 10.32 & 4.60 & 6.68 & 10.26 & 13.96 & 12.50 & 16.14 \\
        NVFP4 & 6.06 & 7.44 & 5.33 & 6.84 & 3.78 & 5.77 & 5.85 & 7.38 & 5.13 & 6.73 & 3.54 & 5.68 & 7.14 & 10.19 & 8.93 & 12.08 \\
        NF4 & 6.04 & 7.51 & 5.40 & 6.89 & 3.79 & 5.81 & 5.93 & 7.48 & 5.21 & 6.77 & 3.60 & 5.71 & 7.40 & 10.54 & 9.35 & 12.54 \\
        Atom & 6.16 & 7.70 & 5.46 & 7.04 & 3.89 & 5.92 & 6.04 & 7.75 & 5.27 & 6.95 & 3.68 & 5.83 & -- & -- & -- & -- \\
        DuQuant & 6.18 & 7.73 & 5.47 & 7.07 & 3.93 & 5.93 & 6.08 & 7.79 & 5.33 & 7.02 & 3.76 & 5.85 & -- & -- & -- & -- \\
        SpinQuant &  6.12 & -- & 5.39 & -- & -- & -- & 5.96 & -- & 5.24 & -- & 3.70 & -- & -- & -- & -- & -- \\
        OSTQuant & 6.07 & -- & 5.40 & -- & -- & -- & 5.91 & -- & 5.25 & -- & 3.59 & -- & -- & -- & -- & -- \\
        4over6 & 6.04 & 7.42 & 5.30 & 6.81 & 3.75 & 5.75 & 5.79 & 7.32 & 5.11 & 6.69 & 3.51 & 5.66 & 7.03 & 10.02 & 8.79 & 11.87 \\
        \rowcolor{red!10} 
        \textbf{\workname{}} & \textbf{5.93} & \textbf{7.32} & \textbf{5.24} & \textbf{6.76} & \textbf{3.71} & \textbf{5.72} & \textbf{5.72} & \textbf{7.25} & \textbf{5.06} & \textbf{6.64} & \textbf{3.47} & \textbf{5.63} & \textbf{6.86} & \textbf{9.78} & \textbf{8.56} & \textbf{11.53} \\
      \Xhline{0.2ex}
    \end{tabular}
  \label{tab:ppl_wakv}
\end{table}

\subsection{Detailed Zero-shot Task Accuracy} \label{apdx:zeroshot_exp}
We provide the detailed zero-shot task accuracy for Llama and Qwen3 models in Table~\ref{tab:zeroshot_llama} and Table~\ref{tab:zeroshot_qwen3}, respectively.

\begin{table} [ht]
  \centering
  \setlength{\tabcolsep}{6pt}
  \renewcommand{\arraystretch}{1.15}
  \footnotesize
  \caption{Detailed zero-shot task accuracy ($\uparrow$) for Llama models under 4-bit weight-activation quantization. KV-cache remains in FP16.}
  \vspace{-5pt}
    \begin{tabular}{ c  c  c c c c c c c }
      \Xhline{0.2ex}
        \textbf{Model} & \textbf{Method} & PIQA & Hellaswag & Winogrande & ARC-e & ARC-c & Lambada & \textbf{Avg.} \\
      \Xhline{0.2ex}\noalign{\vskip 1pt}
 \multirow{6}{*}{Llama-2-7B} & FP16 & 78.45 & 76.17 & 70.24 & 73.82 & 44.97 & 73.28 & 69.49 \\
 \cline{2-9}
 & MXFP4 & 76.61 & 71.31 & 65.27 & 69.15 & 41.38 & 69.69 & 65.57 \\
 & NVFP4 & 77.91 & 75.07 & 68.98 & 71.89 & 43.17 & 72.77 & 68.30 \\
 & MR-GPTQ & 77.86 & 74.99 & 68.82 & 72.94 & 44.37 & 73.14 & 68.69 \\
 & 4over6 & 78.35 & 75.15 & 69.14 & 72.43 & 43.05 & 72.40 & 68.42 \\
 & \cellcolor{red!10}\workname{} & \cellcolor{red!10}78.18 & \cellcolor{red!10}75.16 & \cellcolor{red!10}69.82 & \cellcolor{red!10}72.69 & \cellcolor{red!10}44.49 & \cellcolor{red!10}72.97 & \cellcolor{red!10}\textbf{68.89} \\
      \Xhline{0.2ex}\noalign{\vskip 1pt}
 \multirow{6}{*}{Llama-2-13B} & FP16 & 80.63 & 79.64 & 72.22 & 76.52 & 48.63 & 76.36 & 72.33 \\
 \cline{2-9}
 & MXFP4 & 77.97 & 74.05 & 71.19 & 73.99 & 45.22 & 73.90 & 69.39 \\
 & NVFP4 & 79.92 & 78.54 & 71.19 & 75.51 & 47.27 & 76.05 & 71.41 \\
 & MR-GPTQ & 79.43 & 78.84 & 71.27 & 75.67 & 47.82 & 76.50 & 71.59 \\
 & 4over6 & 79.98 & 78.84 & 71.43 & 75.00 & 46.93 & 75.88 & 71.34 \\
 & \cellcolor{red!10}\workname{} & \cellcolor{red!10}80.20 & \cellcolor{red!10}78.91 & \cellcolor{red!10}73.40 & \cellcolor{red!10}75.80 & \cellcolor{red!10}48.38 & \cellcolor{red!10}76.03 & \cellcolor{red!10}\textbf{72.12} \\
      \Xhline{0.2ex}\noalign{\vskip 1pt}
 \multirow{6}{*}{Llama-3.1-8B} & FP16 & 80.96 & 79.41 & 74.11 & 82.41 & 54.95 & 74.66 & 74.42 \\
 \cline{2-9}
 & MXFP4 & 76.01 & 73.75 & 68.90 & 73.40 & 46.42 & 67.16 & 67.61 \\
 & NVFP4 & 78.67 & 77.98 & 72.22 & 78.54 & 52.56 & 73.63 & 72.27 \\
 & MR-GPTQ & 79.43 & 77.23 & 71.74 & 80.64 & 52.90 & 74.23 & 72.70 \\
 & 4over6 & 79.98 & 78.17 & 72.22 & 80.01 & 54.61 & 73.98 & \textbf{73.16} \\
 & \cellcolor{red!10}\workname{} & \cellcolor{red!10}80.14 & \cellcolor{red!10}77.94 & \cellcolor{red!10}72.38 & \cellcolor{red!10}80.13 & \cellcolor{red!10}53.16 & \cellcolor{red!10}74.13 & \cellcolor{red!10}72.98 \\
      \Xhline{0.2ex}\noalign{\vskip 1pt}
 \multirow{6}{*}{Llama-3.2-3B} & FP16 & 77.69 & 74.08 & 69.53 & 72.05 & 46.16 & 69.71 & 68.20 \\
 \cline{2-9}
 & MXFP4 & 74.59 & 67.76 & 64.96 & 66.04 & 41.04 & 60.08 & 62.41 \\
 & NVFP4 & 75.63 & 71.61 & 67.40 & 69.82 & 44.45 & 67.46 & 66.06 \\
 & MR-GPTQ & 76.93 & 71.68 & 66.54 & 70.16 & 44.80 & 66.99 & 66.18 \\
 & 4over6 & 75.52 & 71.91 & 67.64 & 69.99 & 43.60 & 67.34 & 66.00 \\
 & \cellcolor{red!10}\workname{} & \cellcolor{red!10}76.39 & \cellcolor{red!10}72.43 & \cellcolor{red!10}70.01 & \cellcolor{red!10}69.95 & \cellcolor{red!10}43.00 & \cellcolor{red!10}68.14 & \cellcolor{red!10}\textbf{66.65} \\
      \Xhline{0.2ex}
    \end{tabular}
  \label{tab:zeroshot_llama}
\end{table}

\begin{table} [ht]
  \centering
  \setlength{\tabcolsep}{6pt}
  \renewcommand{\arraystretch}{1.15}
  \footnotesize
  \caption{Detailed zero-shot task accuracy ($\uparrow$) for Qwen3 models under 4-bit weight-activation quantization. KV-cache remains in FP16.}
  \vspace{-5pt}
    \begin{tabular}{ c  c  c c c c c c c }
      \Xhline{0.2ex}
        \textbf{Model} & \textbf{Method} & PIQA & Hellaswag & Winogrande & ARC-e & ARC-c & Lambada & \textbf{Avg.} \\
      \Xhline{0.2ex}\noalign{\vskip 1pt}
 \multirow{6}{*}{Qwen3-4B} & FP16 & 75.03 & 68.54 & 65.67 & 78.32 & 54.10 & 59.36 & 66.84 \\
 \cline{2-9}
 & MXFP4 & 72.80 & 63.40 & 62.27 & 67.89 & 46.84 & 53.56 & 61.13 \\
 & NVFP4 & 73.88 & 66.31 & 61.56 & 73.36 & 50.68 & 53.87 & 63.28 \\
 & MR-GPTQ & 73.39 & 66.11 & 62.59 & 75.13 & 49.91 & 56.18 & 63.89 \\
 & 4over6 & 72.80 & 66.61 & 62.83 & 74.37 & 50.26 & 55.54 & 63.74 \\
 & \cellcolor{red!10}\workname{} & \cellcolor{red!10}73.07 & \cellcolor{red!10}66.52 & \cellcolor{red!10}63.69 & \cellcolor{red!10}74.37 & \cellcolor{red!10}50.94 & \cellcolor{red!10}54.96 & \cellcolor{red!10}\textbf{63.93} \\
      \Xhline{0.2ex}\noalign{\vskip 1pt}
 \multirow{6}{*}{Qwen3-8B} & FP16 & 77.75 & 74.96 & 67.88 & 80.98 & 56.40 & 64.22 & 70.37 \\
 \cline{2-9}
 & MXFP4 & 73.45 & 69.31 & 64.48 & 75.17 & 52.47 & 59.93 & 65.80 \\
 & NVFP4 & 76.22 & 73.29 & 67.96 & 79.88 & 54.95 & 62.62 & 69.15 \\
 & MR-GPTQ & 76.66 & 72.75 & 67.80 & 79.88 & 53.84 & 62.76 & 68.95 \\
 & 4over6 & 75.63 & 73.25 & 67.64 & 78.79 & 54.01 & 62.91 & 68.71 \\
 & \cellcolor{red!10}\workname{} & \cellcolor{red!10}76.55 & \cellcolor{red!10}73.68 & \cellcolor{red!10}68.19 & \cellcolor{red!10}80.56 & \cellcolor{red!10}53.84 & \cellcolor{red!10}62.57 & \cellcolor{red!10}\textbf{69.23} \\
      \Xhline{0.2ex}\noalign{\vskip 1pt}
 \multirow{6}{*}{Qwen3-14B} & FP16 & 79.92 & 78.80 & 72.93 & 82.70 & 60.67 & 67.92 & 73.82 \\
 \cline{2-9}
 & MXFP4 & 77.86 & 75.51 & 69.53 & 79.92 & 56.57 & 63.75 & 70.52 \\
 & NVFP4 & 78.51 & 77.93 & 72.22 & 80.72 & 57.59 & 65.13 & 72.02 \\
 & MR-GPTQ & 79.50 & 77.63 & 71.51 & 80.85 & 58.36 & 66.74 & 72.45 \\
 & 4over6 & 78.29 & 78.52 & 71.59 & 81.48 & 60.24 & 65.11 & 72.54 \\
 & \cellcolor{red!10}\workname{} & \cellcolor{red!10}79.60 & \cellcolor{red!10}78.45 & \cellcolor{red!10}71.82 & \cellcolor{red!10}81.61 & \cellcolor{red!10}59.90 & \cellcolor{red!10}65.77 & \cellcolor{red!10}\textbf{72.86} \\
      \Xhline{0.2ex}\noalign{\vskip 1pt}
 \multirow{6}{*}{Qwen3-32B} & FP16 & 81.94 & 82.59 & 73.40 & 83.25 & 60.84 & 67.13 & 74.86 \\
 \cline{2-9}
 & MXFP4 & 80.14 & 80.43 & 71.35 & 81.99 & 60.41 & 66.39 & 73.45 \\
 & NVFP4 & 79.87 & 81.93 & 69.85 & 82.03 & 61.18 & 66.85 & 73.62 \\
 & MR-GPTQ & 81.66 & 81.90 & 71.51 & 82.53 & 59.90 & 67.18 & 74.11 \\
 & 4over6 & 80.79 & 81.97 & 72.06 & 82.74 & 60.67 & 67.16 & 74.23 \\
 & \cellcolor{red!10}\workname{} & \cellcolor{red!10}80.74 & \cellcolor{red!10}82.17 & \cellcolor{red!10}73.09 & \cellcolor{red!10}82.58 & \cellcolor{red!10}61.60 & \cellcolor{red!10}66.78 & \cellcolor{red!10}\textbf{74.49} \\
      \Xhline{0.2ex}
    \end{tabular}
  \label{tab:zeroshot_qwen3}
\end{table}
\newpage

\section{Additional Kernel Evaluation} \label{apdx:D}

\subsection{Results across Additional Models and Hardware}
\begin{figure}[t]
  \centering
  \includegraphics[width=0.75\linewidth]{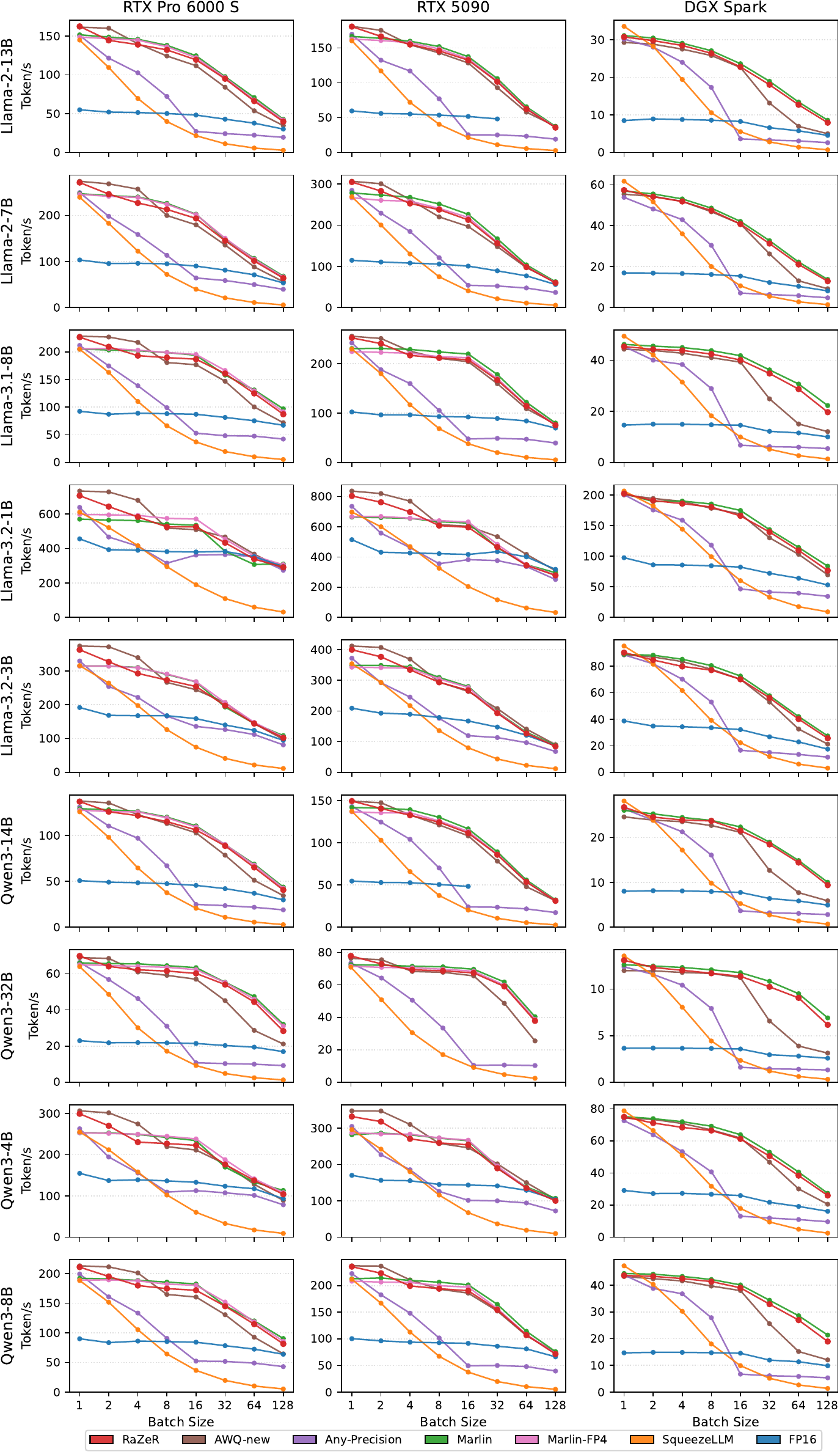}
  \caption{LLM end-to-end decode speed measurements on three GPUs. Token generation speed is reported across batch sizes. Missing data points correspond to out-of-memory configurations.}
  \label{fig:kernel_additional_models}
\end{figure}

We extend the evaluation in the main text to additional models from the Qwen and Llama families with different model sizes and projection shapes, and additionally include results on RTX~5090, which are omitted from the main text for space. \Cref{fig:kernel_additional_models} summarizes end-to-end decoding throughput across eight such models over a range of batch sizes. The observed trends are consistent with the main results: \workname{} achieves near–best-in-class performance at batch size one and maintains competitive throughput at moderate to large batch sizes, closely tracking Marlin and Marlin-FP4 while substantially outperforming AWQ and other weight-only quantized kernels. Results on RTX~5090 follow the same scaling behavior as the server-grade and edge-grade Blackwell GPUs evaluated in the main text, indicating that the performance characteristics of \workname{} are consistent across different classes of Blackwell devices. Some data points are omitted due to out-of-memory (OOM) conditions for certain model and batch-size combinations on specific devices.

\subsection{Weight-Only Kernel Microbenchmarks}
To isolate kernel-level performance from end-to-end model effects, we evaluate weight-only matrix multiplication microbenchmarks across a broad set of matrix shapes and batch sizes. \Cref{tab:kernel_microbench_rtx_pro_6000_s,tab:kernel_microbench_rtx_5090,tab:kernel_microbench_dgx_spark} report kernel execution time for representative $(M,N,K)$ configurations corresponding to the linear projections in Llama~3.1~8B and Qwen3-32B on three GPU platforms.

Across all evaluated configurations, \workname{} achieves performance comparable to or exceeding prior state-of-the-art weight-only quantization kernels. We report results for both CUDA-core– and tensor-core–based \workname{} implementations, which exhibit complementary performance regimes depending on batch size. For each configuration, the faster of the two \workname{} variants is shown in red to indicate the best-performing implementation. Overall, the additional metadata access and remapping logic introduce minimal kernel-level overhead, with \workname{} consistently matching or outperforming existing FP4 and INT4 kernels across models, layers, batch sizes, and GPUs.

\FloatBarrier
\begin{table}[p]
  \centering
  \caption{Kernel latency microbenchmark ($\mu$s, speedup over FP16) (\textsc{RTX Pro 6000 S})}
  \label{tab:kernel_microbench_rtx_pro_6000_s}
  \resizebox{\linewidth}{!}{%
  {\setlength{\tabcolsep}{2pt}%
  \renewcommand{\arraystretch}{1.08}%
  \setlength{\arrayrulewidth}{0.45pt}%
\begin{tabular}{cccccrrrrrrrr}
  \hline
  \textbf{Model} & \textbf{Layer} & \textbf{K} & \textbf{N} & \multicolumn{1}{c|}{\textbf{M}} & \multicolumn{1}{r|}{\textbf{FP16}} & \textbf{RaZeR-CUDA} & \multicolumn{1}{r|}{\textbf{RaZeR-TC}~} & \textbf{Marlin}~~~~~ & \textbf{Marlin-FP4}~ & \textbf{Any-Precision} & \textbf{SqueezeLLM} & \textbf{AWQ}~~~~~~ \\
  \hline
\multirow[c]{32}{*}{Llama-3.1-8B} & \multirow[c]{8}{*}{attn.qkv} & \multirow[c]{8}{*}{4096} & \multirow[c]{8}{*}{6144} & \multicolumn{1}{c|}{\strut 1} & \multicolumn{1}{r|}{\strut 43.1} & \strut \textcolor{red}{12.2 (3.54$\times$)} & \multicolumn{1}{r|}{\strut 18.5 (2.33$\times$)} & \strut 16.9 (2.56$\times$) & \strut 15.0 (2.88$\times$) & \strut 14.7 (2.94$\times$) & \strut 15.7 (2.75$\times$) & \strut 19.1 (2.26$\times$) \\
 &  &  &  & \multicolumn{1}{c|}{\strut 2} & \multicolumn{1}{r|}{\strut 39.6} & \strut \textcolor{red}{16.2 (2.45$\times$)} & \multicolumn{1}{r|}{\strut 18.8 (2.11$\times$)} & \strut 17.2 (2.31$\times$) & \strut 15.2 (2.61$\times$) & \strut 20.2 (1.96$\times$) & \strut 23.0 (1.72$\times$) & \strut 19.3 (2.05$\times$) \\
 &  &  &  & \multicolumn{1}{c|}{\strut 4} & \multicolumn{1}{r|}{\strut 44.3} & \strut 20.5 (2.16$\times$) & \multicolumn{1}{r|}{\strut \textcolor{red}{18.5 (2.39$\times$)}} & \strut 16.9 (2.63$\times$) & \strut 14.9 (2.96$\times$) & \strut 26.7 (1.66$\times$) & \strut 39.5 (1.12$\times$) & \strut 19.5 (2.27$\times$) \\
 &  &  &  & \multicolumn{1}{c|}{\strut 8} & \multicolumn{1}{r|}{\strut 46.8} & \strut 37.2 (1.26$\times$) & \multicolumn{1}{r|}{\strut \textcolor{red}{18.7 (2.51$\times$)}} & \strut 17.1 (2.75$\times$) & \strut 15.1 (3.10$\times$) & \strut 58.1 (0.81$\times$) & \strut 71.6 (0.65$\times$) & \strut 19.4 (2.42$\times$) \\
\hline
  \end{tabular}%
  }%
  }%
\end{table}
\clearpage
\begin{table}[p]
  \centering
  \caption{Kernel latency microbenchmark ($\mu$s, speedup over FP16) (\textsc{RTX 5090})}
  \label{tab:kernel_microbench_rtx_5090}
  \resizebox{\linewidth}{!}{%
  {\setlength{\tabcolsep}{2pt}%
  \renewcommand{\arraystretch}{1.08}%
  \setlength{\arrayrulewidth}{0.45pt}%
  \begin{tabular}{cccccrrrrrrrr}
  \hline
  \textbf{Model} & \textbf{Layer} & \textbf{K} & \textbf{N} & \multicolumn{1}{c|}{\textbf{M}} & \multicolumn{1}{r|}{\textbf{FP16}} & \textbf{RaZeR-CUDA} & \multicolumn{1}{r|}{\textbf{RaZeR-TC}~} & \textbf{Marlin}~~~~~ & \textbf{Marlin-FP4}~ & \textbf{Any-Precision} & \textbf{SqueezeLLM} & \textbf{AWQ}~~~~~~ \\
  \hline
\multirow[c]{32}{*}{Llama-3.1-8B} & \multirow[c]{8}{*}{attn.qkv} & \multirow[c]{8}{*}{4096} & \multirow[c]{8}{*}{6144} & \multicolumn{1}{c|}{\strut 1} & \multicolumn{1}{r|}{\strut 41.1} & \strut \textcolor{red}{12.6 (3.27$\times$)} & \multicolumn{1}{r|}{\strut 18.3 (2.25$\times$)} & \strut 17.2 (2.40$\times$) & \strut 15.9 (2.58$\times$) & \strut 14.2 (2.89$\times$) & \strut 15.5 (2.66$\times$) & \strut 17.8 (2.31$\times$) \\
 &  &  &  & \multicolumn{1}{c|}{\strut 2} & \multicolumn{1}{r|}{\strut 40.9} & \strut \textcolor{red}{17.5 (2.34$\times$)} & \multicolumn{1}{r|}{\strut 18.1 (2.25$\times$)} & \strut 17.3 (2.36$\times$) & \strut 15.9 (2.57$\times$) & \strut 18.6 (2.20$\times$) & \strut 23.3 (1.75$\times$) & \strut 17.8 (2.30$\times$) \\
 &  &  &  & \multicolumn{1}{c|}{\strut 4} & \multicolumn{1}{r|}{\strut 41.2} & \strut 20.6 (2.00$\times$) & \multicolumn{1}{r|}{\strut \textcolor{red}{18.4 (2.24$\times$)}} & \strut 17.1 (2.41$\times$) & \strut 16.0 (2.58$\times$) & \strut 25.2 (1.64$\times$) & \strut 39.6 (1.04$\times$) & \strut 18.0 (2.29$\times$) \\
 &  &  &  & \multicolumn{1}{c|}{\strut 8} & \multicolumn{1}{r|}{\strut 42.4} & \strut 36.0 (1.18$\times$) & \multicolumn{1}{r|}{\strut \textcolor{red}{18.3 (2.31$\times$)}} & \strut 17.8 (2.38$\times$) & \strut 15.9 (2.66$\times$) & \strut 55.6 (0.76$\times$) & \strut 71.6 (0.59$\times$) & \strut 18.0 (2.36$\times$) \\
\hline
  \end{tabular}%
  }%
  }%
\end{table}
\clearpage
\begin{table}[p]
  \centering
  \caption{Kernel latency microbenchmark ($\mu$s, speedup over FP16) (\textsc{DGX Spark})}
  \label{tab:kernel_microbench_dgx_spark}
  \resizebox{\linewidth}{!}{%
  {\setlength{\tabcolsep}{2pt}%
  \renewcommand{\arraystretch}{1.08}%
  \setlength{\arrayrulewidth}{0.45pt}%
  \begin{tabular}{cccccrrrrrrrr}
  \hline
  \textbf{Model} & \textbf{Layer} & \textbf{K} & \textbf{N} & \multicolumn{1}{c|}{\textbf{M}} & \multicolumn{1}{r|}{\textbf{FP16}} & \textbf{RaZeR-CUDA} & \multicolumn{1}{r|}{\textbf{RaZeR-TC}~} & \textbf{Marlin}~~~~~ & \textbf{Marlin-FP4}~ & \textbf{Any-Precision} & \textbf{SqueezeLLM} & \textbf{AWQ}~~~~~~ \\
  \hline
\multirow[c]{32}{*}{Llama-3.1-8B} & \multirow[c]{8}{*}{attn.qkv} & \multirow[c]{8}{*}{4096} & \multirow[c]{8}{*}{6144} & \multicolumn{1}{c|}{\strut 1} & \multicolumn{1}{r|}{\strut 374.7} & \strut 85.2 (4.40$\times$) & \multicolumn{1}{r|}{\strut \textcolor{red}{68.8 (5.44$\times$)}} & \strut 68.6 (5.46$\times$) & \strut 67.3 (5.57$\times$) & \strut 76.5 (4.89$\times$) & \strut 70.8 (5.29$\times$) & \strut 74.5 (5.03$\times$) \\
 &  &  &  & \multicolumn{1}{c|}{\strut 2} & \multicolumn{1}{r|}{\strut 263.8} & \strut \textcolor{red}{80.1 (3.29$\times$)} & \multicolumn{1}{r|}{\strut 83.0 (3.18$\times$)} & \strut 86.2 (3.06$\times$) & \strut 76.8 (3.44$\times$) & \strut 87.6 (3.01$\times$) & \strut 89.7 (2.94$\times$) & \strut 74.6 (3.54$\times$) \\
 &  &  &  & \multicolumn{1}{c|}{\strut 4} & \multicolumn{1}{r|}{\strut 258.0} & \strut 84.0 (3.07$\times$) & \multicolumn{1}{r|}{\strut \textcolor{red}{66.5 (3.88$\times$)}} & \strut 66.0 (3.91$\times$) & \strut 65.5 (3.94$\times$) & \strut 132.6 (1.95$\times$) & \strut 144.7 (1.78$\times$) & \strut 77.3 (3.34$\times$) \\
 &  &  &  & \multicolumn{1}{c|}{\strut 8} & \multicolumn{1}{r|}{\strut 257.5} & \strut 153.8 (1.67$\times$) & \multicolumn{1}{r|}{\strut \textcolor{red}{78.3 (3.29$\times$)}} & \strut 75.1 (3.43$\times$) & \strut 68.1 (3.78$\times$) & \strut 200.6 (1.28$\times$) & \strut 242.1 (1.06$\times$) & \strut 72.9 (3.53$\times$) \\
\hline
  \end{tabular}%
  }%
  }%
\end{table}
\FloatBarrier

\subsection{Tensor-Core Realization of W4A4 \workname{} on Current Hardware}
\label{apdx:w4a4_tensorcore_realization}

Current NVIDIA tensor cores do not expose a native single-pass datapath for executing \workname{} directly in the W4A4 NVFP4 setting. This appendix therefore evaluates an experimental realization of W4A4 \workname{} on existing GPUs, with the goal of understanding the attainable compute throughput on current-generation hardware. We focus on large-batch compute-bound regimes, where weight-only quantization provides limited benefit and can incur additional overhead due to on-the-fly dequantization on the critical compute path. To this end, we implement a \emph{current-hardware realization} using CUTLASS block-scaled NVFP4 GEMM on Blackwell-class GPUs.

This realization decomposes \workname{} into an on-device remapping step followed by two NVFP4 GEMM passes. Due to current hardware constraints, \workname{} is applied to the \emph{weight matrix only}, while activations follow the standard block-scaled NVFP4 tensor-core path.

\paragraph{Two-pass Formulation.}
\workname{} remaps redundant zeros into signed special values using per-group metadata. Because the NVFP4 tensor-core datapath expects standard NVFP4 operands, we realize this remapping via a two-pass decomposition of the weight matrix into two NVFP4 matrices:
\begin{align*}
D &= A B_{\text{main}} + A B_{\text{comp}}.
\end{align*}
The first matrix, $B_{\text{main}}$, replaces each redundant $+0$ with a signed base value, while leaving all nonzero weights unchanged. The second matrix, $B_{\text{comp}}$, provides a small corrective offset so that the sum of the two passes reconstructs the desired special value. Nonzero entries are masked out in $B_{\text{comp}}$.

\vspace{0.25em}
\noindent\textbf{Example (\,$\{\pm5,\pm8\}$\, Configuration).}
For redundant $+0$, we set
\begin{align*}
B_{\text{main}} &: \; +0 \mapsto \pm 4,\\
B_{\text{comp}} &: \; +0 \mapsto
\begin{cases}
\pm 1 & \text{to select } \pm 5,\\
\pm 4 & \text{to select } \pm 8,
\end{cases}
\end{align*}
with the sign and selector determined by 2-bit metadata stored per 16-$K$ group (Sec.~\ref{sec:nvfp4_element_redun}). The two GEMMs are executed back-to-back and accumulated into the same output buffer.

\paragraph{Generality.}
This construction applies to any pair of signed special values that can be expressed as the sum of two FP4-representable values. More generally, for special values $s_0 = x_1 + x_2$ and $s_1 = y_1 + y_2$ with FP4-representable $x_1, x_2, y_1, y_2$, $B_{\text{main}}$ injects the first component while $B_{\text{comp}}$ injects the second according to the per-group selector bit. All nonzero weights are preserved in $B_{\text{main}}$ and masked out in $B_{\text{comp}}$. Beyond the values already representable in FP4, this construction additionally supports the signed special values $\{\pm 2.5, \pm 3.5, \pm 4.5, \pm 5, \pm 5.5, \pm 6.5, \pm 7, \pm 7.5, \pm 8, \pm 9, \pm 10, \pm 12\}$.

\paragraph{Implementation Details.}
To avoid host-side unpacking and repacking, we generate $B_{\text{main}}$ and $B_{\text{comp}}$ directly on the GPU in the packed sub-byte layout consumed by CUTLASS. Both passes use block-scaled NVFP4 GEMM with floating-point accumulation and output, enabling correct two-pass accumulation without introducing additional quantization error.

\begin{figure}[H]
  \centering
  \includegraphics[width=0.88\linewidth]{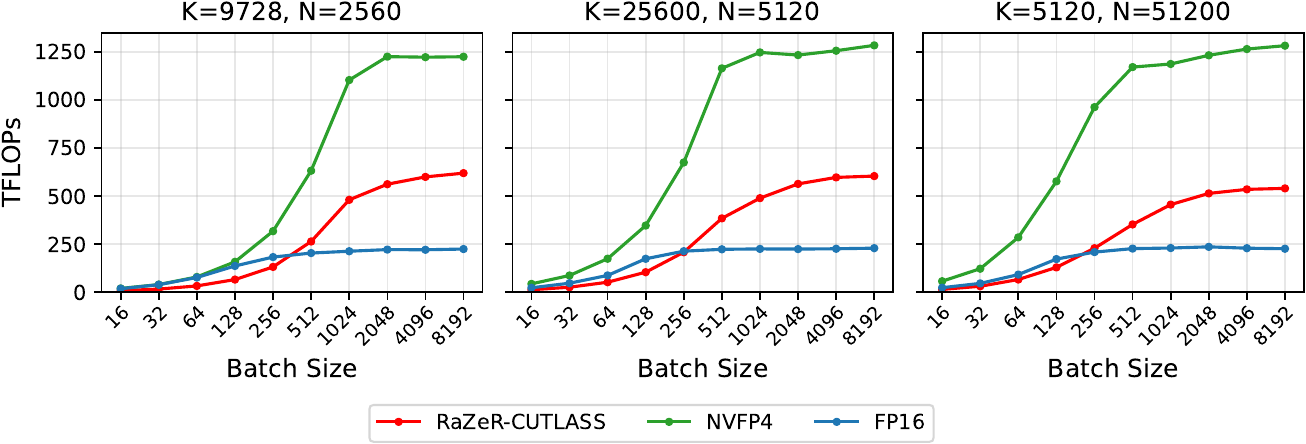}
  \caption{Throughput of W4A4 \workname{} implemented using a two-pass block-scaled NVFP4 GEMM on an RTX~5090. Throughput is computed from the execution time of the two GEMM passes and normalized to an effective operation count of $2MNK$. cuBLAS FP16 and native NVFP4 GEMM baselines are shown for reference.}
  \label{fig:w4a4_nvfp4_tflops}
\end{figure}

\paragraph{Performance.}
Figure~\ref{fig:w4a4_nvfp4_tflops} reports the throughput of the two-pass W4A4 \workname{} realization across batch sizes and matrix shapes on RTX~5090. Throughput is reported using an effective operation count of $2MNK$ multiply-adds, accounting for both GEMM passes. Across all configurations, throughput increases with batch size and saturates in the compute-bound regime. In this regime, the two-pass NVFP4 \workname{} realization consistently delivers more than $2\times$ higher throughput over FP16 GEMM.

Despite these gains, performance remains below that of native single-pass NVFP4 GEMM, reflecting the unavoidable overhead of executing two separate passes. In particular, the compensation matrix $B_{\text{comp}}$ is inherently sparse, as it contains nonzero entries only at locations corresponding to redundant zeros. In our current implementation, this sparsity is not exploited, and leveraging it within a two-pass realization (e.g., by reducing redundant computation or improving pass fusion) remains an interesting direction for future work. These observations further motivate the custom \workname{} tensor-core architecture proposed in Sec.~\ref{sec:nvfp4_mac}, which eliminates the need for multi-pass execution altogether.

\newpage

\section{Additional Experiments on Auto-tuned Kernel Performance} \label{apdx:E}

\begin{figure}[H]
  \centering
  \includegraphics[width=0.8\linewidth]{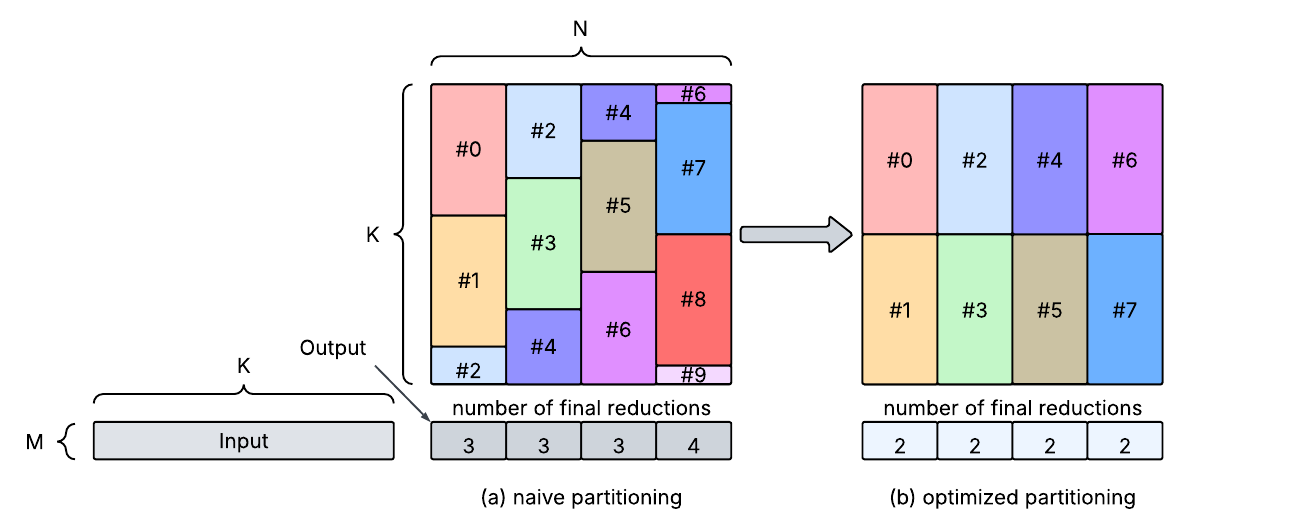}
  \caption{Optimized partitioning for small weight tensors. The optimized strategy reduces the number of global reduction stages and associated overhead.}
  \label{fig:optimized-partition}
\end{figure}

\vspace{-12pt}

In the \workname{} weight-only tensor-core kernel, the weight matrix is partitioned into multiple stripes of approximately equal length, potentially spanning multiple $N$-dimensions. Each stripe's length must be a multiple of 256 to satisfy low-level kernel design constraints. Each thread block fully occupies one SM, processes one stripe, and accumulates partial results to global memory.

On modern GPUs with a large number of SMs, this strategy can excessively partition small matrices. We observe that in the memory-bound scenario, reducing the number of active SMs does not underutilize memory bandwidth. Motivated by this observation, we introduce an auto-tuning strategy that reduces the number of partitions and global reduction stages to minimize GEMM and GEMV latency.

\Cref{fig:optimized-partition} illustrates the approach. For a GPU with 10 SMs, the naive strategy utilizes all SMs, resulting in 3--4 reduction stages per output tile. In contrast, the optimized strategy launches only 8 SMs, reducing the number of global reductions and improving GEMM/GEMV latency. Table~\ref{tab:autotune} reports profiling results on an NVIDIA RTX 5090 (170 SMs), where the auto-tuned kernel achieves up to 9.87\% throughput improvement.

\begin{table}[H]
\centering
\caption{End-to-end inference decode speed measurements on RTX 5090 for several small models..}
\label{tab:autotune}
\resizebox{0.5\textwidth}{!}{

\begin{tabular}{l|l|ll|r}
\hline
Model                         & \begin{tabular}[c]{@{}l@{}}Batch\\ Size\end{tabular} & \begin{tabular}[c]{@{}l@{}}Default\\ (tok/s)\end{tabular} & \begin{tabular}[c]{@{}l@{}}Auto-tuned\\ (tok/s)\end{tabular} & \multicolumn{1}{l}{Improvement} \\ \hline
\multirow{7}{*}{Llama-3.2-1B} & 1                                                    & 611.04                                                      & 653.83                                                       & 7.00\%                          \\
                              & 2                                                    & 610.40                                                      & 652.23                                                       & 6.85\%                          \\
                              & 4                                                    & 604.65                                                      & 644.40                                                       & 6.57\%                          \\
                              & 8                                                    & 603.85                                                      & 628.73                                                       & 4.12\%                          \\
                              & 16                                                   & 578.74                                                      & 620.68                                                       & 7.25\%                          \\
                              & 32                                                   & 430.81                                                      & 473.31                                                       & 9.87\%                          \\
                              & 64                                                   & 333.40                                                      & 365.10                                                       & 9.51\%                          \\ \hline
\multirow{7}{*}{Llama-3.2-3B} & 1                                                    & 329.40                                                      & 338.95                                                       & 2.90\%                          \\
                              & 2                                                    & 323.05                                                      & 337.90                                                       & 4.60\%                          \\
                              & 4                                                    & 319.45                                                      & 340.25                                                       & 6.51\%                          \\
                              & 8                                                    & 296.94                                                      & 317.18                                                       & 6.82\%                          \\
                              & 16                                                   & 268.69                                                      & 274.85                                                       & 2.29\%                          \\
                              & 32                                                   & 190.57                                                      & 199.68                                                       & 4.78\%                          \\
                              & 64                                                   & 128.70                                                      & 129.39                                                       & 0.54\%                          \\ \hline
\multirow{7}{*}{Llama-3.1-8B} & 1                                                    & 222.133                                                     & 233.18                                                       & 4.97\%                          \\
                              & 2                                                    & 216.79                                                      & 222.06                                                       & 2.43\%                          \\
                              & 4                                                    & 215.46                                                      & 227.00                                                       & 5.36\%                          \\
                              & 8                                                    & 209.43                                                      & 225.67                                                       & 7.75\%                          \\
                              & 16                                                   & 207.24                                                      & 219.93                                                       & 6.12\%                          \\
                              & 32                                                   & 170.03                                                      & 170.70                                                       & 0.39\%                          \\
                              & 64                                                   & 117.15                                                      & 118.24                                                       & 0.93\%                          \\ \hline
\end{tabular}
}
\end{table}


\end{document}